\documentclass[letterpaper, 10 pt, conference]{ieeeconf}
\IEEEoverridecommandlockouts      
\PassOptionsToPackage{hyphens}{url}\usepackage{hyperref}

\usepackage{graphics} 
\usepackage{graphicx} %

\usepackage{tikz}
\usetikzlibrary{calc}

\usepackage{amsmath} 
\usepackage{amssymb}  
\usepackage{mathtools}

\DeclareMathOperator{\dist}{d}
\DeclareMathOperator{\proj}{Proj}
\DeclareMathOperator{\atan}{atan}

\usepackage[ruled, vlined]{algorithm2e}
\let\labelindent\relax
\usepackage{enumitem}
\usepackage{multicol}
\usepackage{multirow}

\usepackage{cite}

\graphicspath{ {../figs/}{../figures/}}

\newcommand{\algName}{potential gap}

\title{\LARGE \bf Potential Gap: Using Reactive Policies to Guarantee
Safe Navigation}

\author{{Ruoyang Xu$^{1,\dagger}$, Shiyu Feng$^{2,\dagger}$, and Patricio A. Vela$^{1}$}
\thanks{$^{1}$ R. Xu was, and P.A. Vela is with the School of Electrical and Computer
Engineering and the Institute for Robotics and Intelligent Machines, Georgia Institute of Technology, Atlanta, GA 30308, USA.
{\tt\small \{rxu74, pvela\}@gatech.edu}}%
\thanks{$^{2}$S. Feng is with the School of Mechanical Engineering and the School of Electrical and Computer Engineering, Georgia Institute of Technology, Atlanta, GA 30308, USA.
{\tt\small shiyufeng@gatech.edu}}
\thanks{$\dagger$ Equal contribution}
\thanks{*This work supported in part by NSF Award \#1849333.}
}

\begin{document}

\maketitle
\thispagestyle{empty}
\pagestyle{empty}

\begin{abstract}
This paper considers the integration of gap-based local navigation
methods with artificial potential field (APF) methods to derive a local
planning module for hierarchical navigation systems that has provable
collision-free properties. Given that APF theory applies to idealized
robot models, the provable properties are lost when applied to more
realistic models. We describe a set of algorithm modifications that
correct for these errors and enhance robustness to non-ideal models. 
Central to the construction of the local planner is the use of
sensory-derived local free-space models that detect gaps and use
them for the synthesis of the APF. Modifications are given for a
nonholonomic robot model. Integration of the local planner, called {\em
potential gap}, into a hierarchical navigation system provides the local
goals and trajectories needed for collision-free navigation through
unknown environments.
Monte Carlo experiments in benchmark worlds confirm the asserted safety
and robustness properties by testing under various robot models.
\end{abstract}


\section{Introduction}

Potential methods generate guaranteed safe reactive policies for fully
controlled point-mass agents going from one point in space
to another, under the assumption of known collision space regions
\cite{APF1985}. 
A non-trivial performance gap emerges between idealized and actual
models, the latter of which 
occupy space, 
have dynamics, 
can be nonholonomic, and 
have incomplete collision space measurements (from a limited
field of view) \cite{APF_limitation, Sensor_nonholo} .
A variety of other planning schemes have been derived and refined to
address the performance gap 
\cite{Smith2020, APF_feedback_diff}. 
In the context of long distance navigation, no single strategy provides
the best outcome.  Hierarchical strategies permit multiple approaches
that operate at different temporal and spatial scales
\cite{Smith2020, HierLatombe, HierByrne}. 
More importantly, they consider the role of visual sensors in
providing up-to-date information regarding navigable space that
complements existing and potentially stale free space maps.
Unfortunately, more research effort goes into the design of
general-purpose, single strategy planning schemes, rather than into the
closed-loop performance of sensing-informed hierarchical navigation
methods.

The research deficit implies that there may be value in more extensive
studies of hierarchical methods and how classical navigation approaches 
may successfully integrate into their architecture. 
Doing so may also suggest how to best leverage general-purpose planners
within an equally general and modular hierarchical navigation framework.
It also provides an excellent opportunity to reconsider how idealized
approaches, or those with theoretical support but limited applicability,
might provide safe local planners that operate on short temporal and
small spatial scales.  {\em For these smaller domains, the 
fragile theoretical guarantees could translate more robustly to
non-ideal settings through targeted modifications}.  
This paper explores the
value of reconsidering potential field navigation strategies for
vision-based navigation through unknown environments. In the process, it
creates a hierarchy of modular components that are activated as the
robot model strays from the ideal one. Each component corrects for the
loss of theoretical support and minimizes sensitivity to this loss.
The final approach, called the {\em potential gap} local planner, is
shown to inherit the navigation guarantees of potential field methods
for idealized systems, to demonstrate robustness to non-ideal robot
models, and to degrade gracefully for uncorrected non-ideal properties.
It provides a framework for improved safety within a
real-time, hierarchical navigation system.

\subsection{Related Work and Research Context \label{sec:BG}}

Visual navigation has an extensive history that bridges several core
areas of robotics from perception to planning to action. This section
narrowly reviews topics most related to the study at hand, starting at
the top level with hierarchical navigation and descending to closed-loop
execution.

\subsubsection{Hierarchical Navigation Systems} \label{sss:HierNavSys}
Navigation systems for mobile vehicles must identify a solution to the
high-level task of goal attainment with realizable low-level system
control laws.
In the context of mobile robot navigation, bipartite hierarchical
systems
often consist of a global
planner that  generates paths with prior knowledge of the environment, 
and a local planner that reacts to the local information obtained from
the real-time sensors \cite{Smith2020}.  They leverage the advantages
of different planners while offsetting their limitations, and have long
been used in robotics, as the hierarchical structure leads to higher
fault tolerance and increased robustness 
\cite{6722516,1227193,407602,5547041}.
The global planner, which runs at a lower rate and considers all known
map information to provide a global path from a given start pose to a
goal pose, can be any applicable planner \cite{lavalle2006planning}.
In contrast, the local planner executes at a higher frequency to produce
kinodynamically controllable commands for driving robots and avoiding
obstacles. There is flexibility to select the local planner so long as
it is structured to receive sensory input and assimilate the input into
its local collision-avoiding navigation decision structure
\cite{VFH,DWA,TEB,SmEtAl[2020]egoTEB}. Its limited scale supports
real-time, sensory-driven navigation with collision avoiding properties.

\subsubsection{Reactive and Potential Field Methods}
The artificial potential field method and its variants are a family of
planners whose instances offer simple and fast computation for mobile
robot obstacle avoidance 
\cite{APF1985,APF_numerical,APF_newfunction, APFSimAnnel, APFRegression}. 
While the potential field is particularly attractive due to its elegance
and simplicity, there are substantial shortcomings inherent to this
method such as a lack of consideration for robot kinematics, dynamics,
as well as local minimum problems with regard to world geometry
\cite{APF_limitation, APF_feedback_diff}. 
Significant efforts were made to alleviate those
problems \cite{APF_guarantee, APFLaplace}. 
When implemented as reactive planners, the family of APFs nevertheless
directly map robot state and sensor observation to available actions,
offering better computational performance than deliberative planners. 
They share valuable traits with perception space methods,
such as minimal sensor processing and planning complexity in the
egocentric robot frame. Integrating reactive methods and perception
space methods can leverage their fast compute properties and the
limited deliberation associated with local planning modules informed by
a global planner.



\subsubsection{Perception-Space and Gap Navigation Methods}
Recent work has explored the use of perception-space representations,
inspired from Marr's 2.5D space, and argued in favor of mixed
representation hierarchical navigation strategies
\cite{Smith2020,SmEtAl[2020]egoTEB}. 
In particular, a local planner (limited to a short time and small
spatial scale) gains computational advantages by minimally processing
the sensor data and recasting local navigation as an ego-centric
decision process.  Related work has established similar benefits for
stereo MAVs \cite{Matthies2014StereoVO}.  

For ground vehicle navigation, the topology of the space and how to
interpret it from sensor data has emerged as an important local navigation
decision \cite{SmEtAl[2020]egoTEB}. Gap-based processing aimed at
detecting passable free-space is compatible with and improves the
synthesis of local optimal paths.
While there is no widely agreed-upon representation--or processing
approach--for gaps, gap-based methods typically reduce 1D laser scan
measurements to a set of ``gaps'' comprised of beginning and terminating
points that represent collision-free regions in the observable space
\cite{followthegap, CG, TCG, SG, AG}.
A gap-based method would then generate reactive motion commands towards a
selected gap, which gives favorable performance in reasonably sparse
and structured environments.
However existing gap representations either rely on the presence of free
space to infinity \cite{followthegap} or dense and accurate sensor
measurements for detecting discontinuities \cite{CG}, leaving
the representation prone to potentially false collision free spaces and errors.
Gap-based methods operate
semi-deliberately in that the construction and selection of free space
are intentional while execution is reactive. 
There is an opportunity to connect perception space methods with
classical potential field methods through the affordance of gaps to
deliver safe planning.

The value of gap detection in terms of establishing {\em navigation
affordances} for local planning was established in \cite{SmEtAl[2020]egoTEB}.
The intent of this study is to explore more deeply the gap
representations associated with gap-based path planning and their
connection to safe navigation through local space.  Based on their use,
gap-based methods will naturally connect the local planning module to
the potential field navigation strategies reviewed above, in turn
overcoming the latter's fundamental limitations for complex worlds.




\subsubsection{Safety in Control}
Control barrier functions (CBFs) are a class of functions whose role is
to serve as instantaneous, point-wise constraints on the applied control
to achieve forward invariance with respect to a safe set, in our context
a {\em collision-free set} (or to achieve forward reachability then
invariance with respect to a {\em target set}) \cite{AmesCBFReview}.
By converting safety (reachability) constraints into
linear-in-the-control constraint equations, CBFs simplify the
certification of safety (reachability) with a trade-off in
short-horizon optimality.  Looking back at the history of CBFs, which
derive from barrier functions, there is a connection to soft-constraint
optimal control methods.  Barriers were used to generate
soft-constraint modifications to the optimization landscape, and could
be used to generate a family of optimal control solutions from a
parametrically varied barrier specification, which often had properties
similar to or were potential fields. By sequentially solving and
warm-starting a series of soft-constraint problems with increasing
penalties, the intent was to converge to the equivalent hard-constraint
solution.  In connecting potential methods to hierarchical planning, we
have the opportunity to translate the collision-critical potentials to
CBF-like specifications for ensuring the safe execution of local paths.

For an idealized robot model (full control, 1st order, point-mass, etc.), 
the safe forward-invariant set from a CBF specification aligns with the
safe reachable set generated using HJB methods \cite{HJTomlin}. 
This set is described by a barrier function level-set. For non-ideal
robots a gap emerges between the idealized level-set defined by the
barrier function and the safe set defined by HJB reachability/CBF
feasibility analysis.  Both approaches require offline analysis to
characterize the gap to arrive at the correct set.  This paper
takes a step towards efficiently and online modifying an idealized,
sensor-derived safe set and applied controls for non-ideal robots.

\subsection{Hierarchical Planning for Navigation}

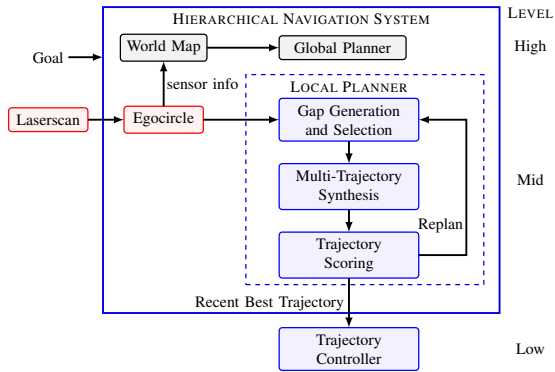
\begin{figure}[t]
  \vspace*{0.07in}

  \hspace*{-0.2in}
  \scalebox{0.6}{\tikzstyle{block} = [draw, rectangle, text centered, thick,rounded corners=2pt,
                     minimum height=1.5em, minimum width=5em, inner sep=4pt]
\tikzstyle{typical} = [fill=white!95!black]
\tikzstyle{reddish} = [draw=red,fill=white!95!red]
\tikzstyle{blueish} = [draw=blue,fill=white!95!blue]
\tikzstyle{greenish} = [draw=green!40!gray,fill=white!95!green]
\tikzstyle{longblock} = [draw,rectangle,text centered,thick,rounded corners=2pt,
                     minimum height=1.5em, minimum width=8em, inner sep=4pt]
\tikzstyle{largeBlock} = [draw, rectangle, very thick,
                     minimum height=19.3em, minimum width=25em, inner sep=4pt]
\tikzstyle{smallBlock} = [draw, rectangle, text centered, thick, dashed,
                     minimum height=13.25em, minimum width=15em, inner sep=4pt]
\tikzstyle{dashedBlock} = [draw, dashed, rectangle,
                     minimum height=2em, minimum width=4em, inner sep=4pt]
\tikzstyle{newtip} = [->, very thick]
\tikzstyle{bidir} = [<->, very thick]
\tikzstyle{newtip_dashed} = [->, very thick, dashed]
\begin{tikzpicture}[auto, inner sep=0pt, outer sep=0pt, >=latex]

  \node[block, reddish] (laserscan) {Laserscan};

  \node[anchor=south] (goal) at ($(laserscan.north) + (0, 1.0)$) {Goal};
    
  \node[block,reddish,anchor=west] (egocircle) 
    at ($(laserscan.east) + (0.80, 0)$) 
    {\centering Egocircle};
  \node[block, typical, anchor=south] (worldmap)  
    at ($(egocircle.north)+(0, 1)$)
    {\centering World Map};

  \node[longblock, typical, anchor=west] (global) 
    at ($(worldmap.east) + (1.6, 0)$)  
    {\centering Global Planner};

  \node[longblock, blueish, anchor=west, text width=8em] 
    (gap) at (egocircle-|global.west)
    {\centering Gap Generation \\ and Selection};

  \node[longblock,blueish,anchor=north, text width=8em] (traj) 
    at ($(gap.south) - (0, 0.5)$) 
    {\centering Multi-Trajectory \\ Synthesis};
  
  \node[longblock,blueish,anchor=north, text width=8em] (score) 
    at ($(traj.south) - (0, 0.5)$) {\centering Trajectory \\ Scoring};
    
  \node[longblock,blueish,anchor=north, text width=8em] (controller) 
    at ($(score.south) - (0, 1.1)$) {\centering Trajectory \\ Controller};
    
  \node[smallBlock, draw=blue!90!black,anchor=north] (local) 
    at ($(gap)!0.09!(gap) + (1em, 1)$){};
  \node[anchor=north,xshift=-1em,yshift=-4pt] (localtext) at (local.north) 
      {\sc Local Planner};

  \node[largeBlock, draw=blue!90!black,anchor=north] (planning) 
    at ($(global)!0.09!(worldmap) + (-0.9-1.5em, 0.9)$){};
  \node[anchor=north,xshift=0em,yshift=-4pt] (planningtext) at (planning.north) 
      {\sc Hierarchical Navigation System};

  \draw[newtip] (laserscan.east) -- (egocircle.west);
  \draw[newtip] (egocircle.north) -- node[midway,right,xshift=2pt]{sensor info} 
    (worldmap.south);

  \draw[newtip] (worldmap.east) -- (global.west);
  \draw[newtip] (egocircle.east) -- (gap.west);

  \draw[newtip] (gap.south) -- (traj.north);
  \draw[newtip] (traj.south) -- (score.north);

  \draw[newtip] (score.south) -- node [midway,left,xshift=-3pt]{Recent Best Trajectory}(controller.north);
  
  \draw[newtip,<-] (gap.east) -- ($(gap.east)+(30pt,0pt)$) |-
  node[midway,left,anchor=east,xshift=-2pt,yshift=19pt]{Replan}(score.east);

  \draw[newtip] ($(goal.east)+(3pt,0pt)$) -- (goal.east-|planning.west);
%
%

  \node[anchor=north west, xshift=5pt] (LT) at (planning.north east)
   {\sc Level};
  \node[anchor=center] at (LT|-global) {High};
  \node[anchor=center] at (LT|-local) {Mid};
  \node[anchor=center] at (LT|-controller) {Low};
\end{tikzpicture}}
  \caption{Hierarchical Navigation System information flow with global
  planner modules (gray boxes) and local planner modules (blue boxes).}
  \label{fig:hie_chart}
\end{figure}

As briefly noted in \S \ref{sss:HierNavSys} and depicted in Figure
\ref{fig:hie_chart}, a hierarchical navigation system decomposes its task
into two parts with distinct temporal and spatial scales: a global
planner and a local planner.  Given a goal point and an estimated map, the
global planner synthesizes a candidate path from the current robot pose
to the goal.  The global path is passed to the local planner, whose
objective is to generate local goals and local paths that lead the robot
to the final goal.  The reduced temporal and spatial scale of the local
planner permits real-time operation in the face of novel information
generated from sensor measurements. These measurements are also passed to
the global planner to enrich the global map for generating valid global paths. 
The objective of the local planner, as informed by
the global path, is to sequentially synthesize collision-free
trajectory segments whose concatenation terminates at the
final goal.  The {\em potential gap} method described next 
will define the operation of the local planner.

%


\section{The Potential Gap Local Navigation Module} \label{sec:PotentialGap}

This section describes the approach taken to identify {\em gaps} for
navigation purposes and how they are used to specify potential-based local
navigation vector fields. By design, the vector fields have guaranteed
collision-free {\em passage through the gap} for idealized settings
(first-order, point-mass, holonomic travel, full vision). For
non-idealized robot models, the potential fields require modification
to their construction and use as barrier functions for an additional
run-time hedge against collisions. 

%

\newcommand{\closeGap}{Close Gap}
\newcommand{\gapSet}{\mathcal{G}}
\newcommand{\thAlpha}{\tau_\alpha}
\newcommand{\agP}{\epsilon}

\newcommand{\xLG}{x^*_{\text LG}}
\newcommand{\rgpSet}{\mathcal{G}_{\text{SGP}}}
\newcommand{\attPot}{\Phi}
\newcommand{\rotVF}{\Theta}
\newcommand{\J}{\mathbb{J}}
\newcommand{\pLeft}{p_l}
\newcommand{\pRight}{p_r}
\newcommand{\gap}{G}
\newcommand{\gapS}{G'}
\newcommand{\threshGA}{\tau_{\text{GA}}}
\newcommand{\bandGA}{\sigma}
\newcommand{\rIns}{r_{\text{ins}}}

\subsection{Gap Analysis and Gap Detection} \label{sec:gap-detect}

Existing gap detection methods considered a limited set of navigation and
sensing scenarios, leading to correct but not fully generalizable gap
detections. Though {\closeGap} sought to address these circumstances, the
set of scenarios covered was still limited \cite{CG}.
Our conception of a {\em gap} is informed by earlier work using gaps to
define local navigation goals \cite{SmEtAl[2020]egoTEB}. 
In considering the intent behind gap-based methods and their utility for
navigation, we propose a different conception of gaps that leads to a new
method for detection and analysis. 

Viewing gaps from a perception space approach, it is equally natural to
represent them in robot-centered polar coordinates. This perspective leads
to the categorization of gaps into swept and radial gaps based
on their dominant direction.
Conceptually, swept gaps involve curves with larger angular sweeps,
while radial gaps have larger distance variation and little angular difference.
The former ``face" the robot and have good line-of-sight visibility
properties.  Swept gaps provide information both about passage {\em to}
the gap, as well as {\em through} the gap.  Radial gaps are oriented
``sideways" to the robot and have bad line-of-sight visibility
properties.  They exhibit poor properties regarding passage to the gap
and/or through it, see Fig.~\ref{fig:GapGeom}(b) and 
Fig.~\ref{fig:gap-parsing-comparison} A. The objective of a gap detection
strategy should be to provide the maximal quantity of swept gaps to
leverage their favorable properties.


\subsubsection{Swept and Radial Gap Detection}
We define a new gap detection policy that simplifies gaps when they
arise from complex world structure. 
The policy avoids potentially false information arising from measurement
sparsity.  Gap modeling hinges on classifying them as swept or radial.

\begin{figure}[t]
  \vspace*{0.07in}
  \begin{tikzpicture}[inner sep=0pt, outer sep=0pt]
    \node [anchor= south west] (SWEPTG) at (0.05, 0.2)
    {{\includegraphics[width=0.25\columnwidth]{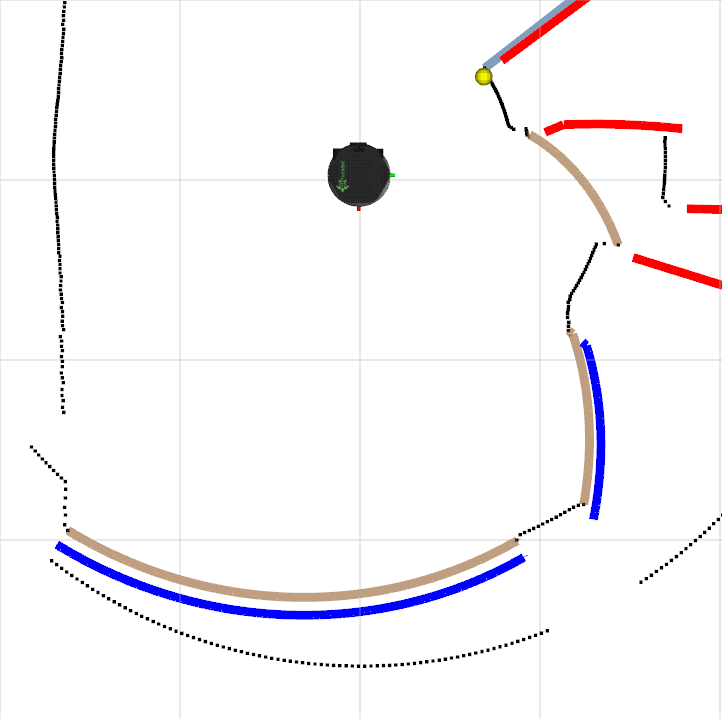}}};
    \node [anchor=south west] (RADIALG1) at (0.05 + 0.25\columnwidth, 0.2)
    {{\includegraphics[width=0.25\columnwidth]{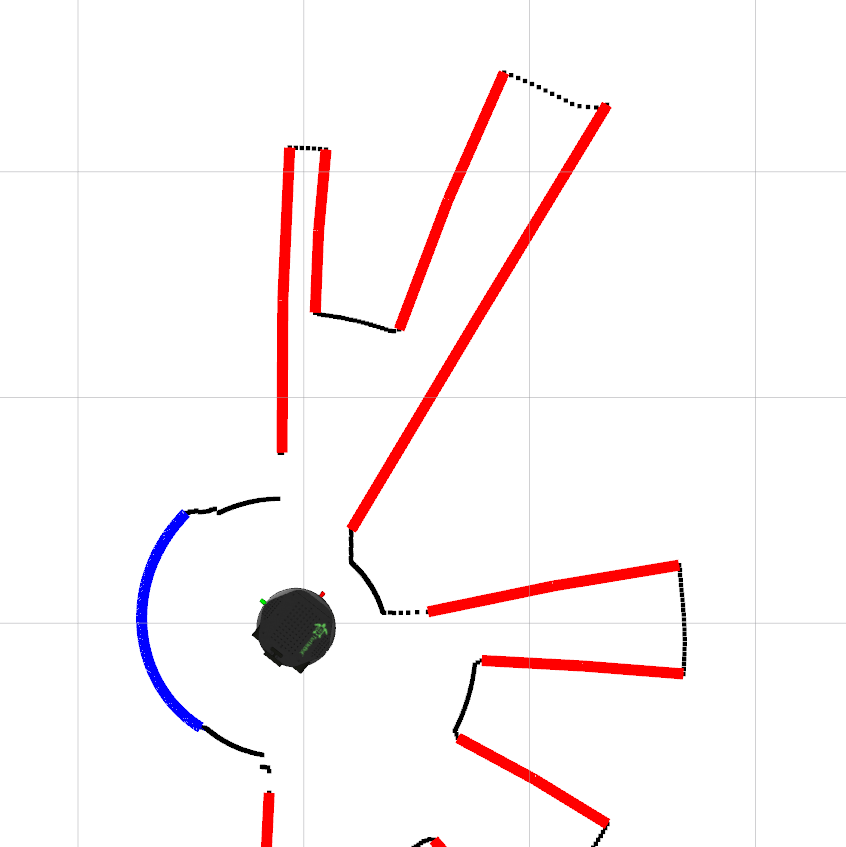}}};
    \node [anchor=south west] (RADIALG2) at (0.05 + 0.5\columnwidth, 0.2)
    {{\includegraphics[width=0.25\columnwidth]{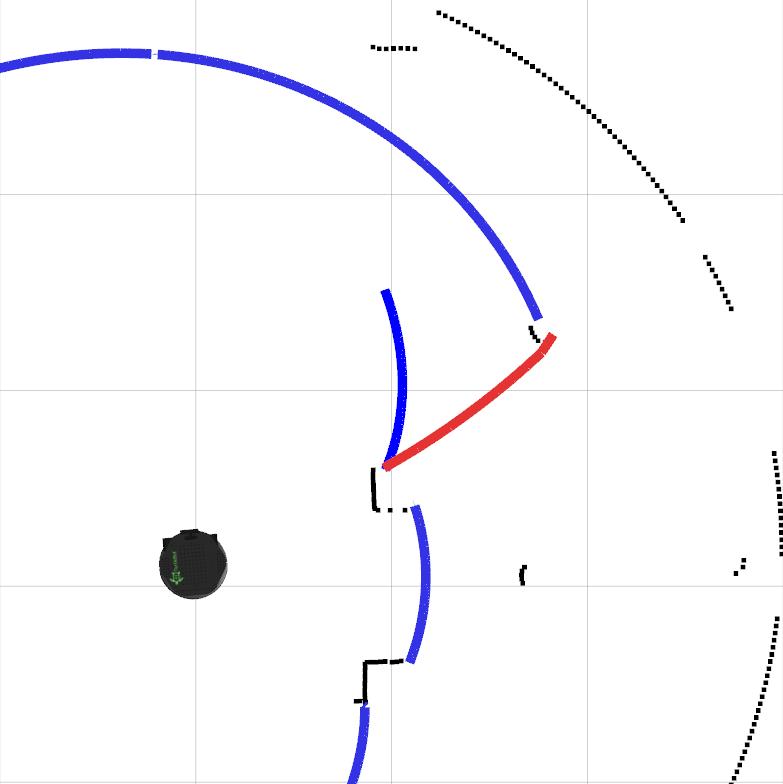}}};
    \node [anchor=south west] (RADIALG3) at (0.05 + 0.75\columnwidth, 0.2)
    {{\includegraphics[width=0.25\columnwidth]{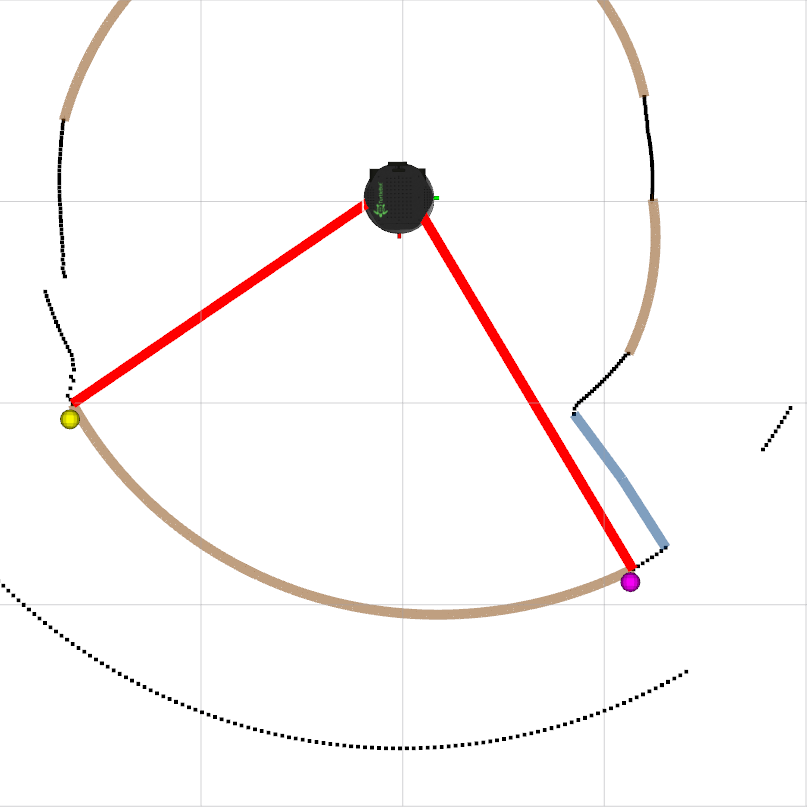}}};


    \node[anchor=south west] (f4gl) at (7, 1.7) {\textcolor{black}{$l_l$}};
    \node[anchor=south west] (f4gr) at (8, 1.3) {\textcolor{black}{$l_r$}};
    \draw[red, thin] (6.65, 1.3) -- (8.2, 0.8);
    \node[anchor=south west] (f4alpha) at (6.85, 1.25) {\textcolor{black}{$\alpha$}};

    \node[anchor=south west] at (SWEPTG.south west) {\scriptsize (a)};
    \node[anchor=south west] at (RADIALG1.south west) {\scriptsize (b)};
    \node[anchor=south west] at (RADIALG2.south west) {\scriptsize (c)};
    \node[anchor=south west] at (RADIALG3.south west) {\scriptsize (d)};

    \node[anchor=south,xshift=5pt,yshift=-5pt] (tsg0) at (SWEPTG.south) 
        {\scriptsize swept gap};
    \node[yshift=1pt] (tsg) at (tsg0.north) {};

    \draw[->] (tsg.north) to[out=140,in=290] ($(tsg.north)+(-5pt,6pt)$);
    \draw[->] (tsg.north) to[out=90,in=270] ($(tsg.north)+(15pt,14pt)$);
    \draw[->] (tsg.north) to[out=40,in=200] ($(tsg.north)+(33pt,12pt)$);

    \node[anchor=north] (trg) at (RADIALG1.120) {\scriptsize radial gap};

    \draw[->] (trg.south) to[out=270,in=0] ($(trg.south)+(-10pt,-6pt)$);
    \draw[->] (trg.south) to[out=270,in=150] ($(trg.south)+( 6pt,-9pt)$);
    \draw[->] (trg.south east) to ($(trg.south east)+( 5pt,-2pt)$);

    \node[anchor=north,xshift=-3pt,yshift=-30pt] (srg) at (SWEPTG.north) 
      {\scriptsize simplified gap};
    \draw[->] ($(srg.north)+(0,2pt)$) 
        to[out=90,in=225] ($(srg.north)+(21pt,12pt)$);

    \node[anchor=north east] (rgc) at ($(RADIALG2.north) + (0pt, 1pt)$)
      {\scriptsize radial gap};
    \node[anchor=north west, yshift=-0.1em, fill=white] (to) at (rgc.south west)
      {\scriptsize to};
    \node[anchor=north west, yshift=-0.1em, fill=white] (sgc) at (to.south west)
      {\scriptsize swept gap};

    \draw[->] (sgc.200) to[out=270,in=200] ($(sgc.south)+(10pt,-10pt)$);
    \draw[->] ($(rgc.east)+(1pt,0pt)$) to[out=0,in=60] ($(rgc.east)+(13pt,-20pt)$);

  \end{tikzpicture}
  \caption{(a,b) Visual of swept and radial gaps, (c) the radial gap
  conversion process, and (d) radial gap angular geometry. Black dot
  denotes the robot. \label{fig:GapGeom}}
\end{figure}
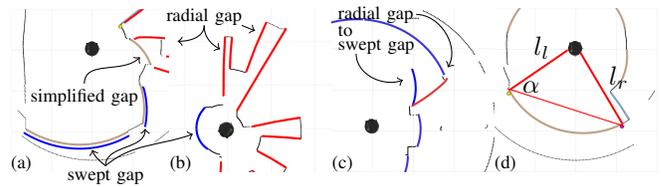

Suppose that a full (360$^\circ$) laser scan-like measurement $\mathcal{L}$
is available to the robot. In the instance of a limited field of view, 
measurement propagation using the \emph{egocircle} \cite{SmEtAl[2020]egoTEB} 
would keep track of measured parts of the surrounding collision-space.
Let the polar measurement $\mathcal{L}$ provide $n$ readings with
a maximum range value $d_{\max}$.
Perform a first pass through $\mathcal{L}$ looking for the following
two outcomes:
\begin{enumerate}[leftmargin=*]
  \item Large interval $I = [i,i+k] \ni
    \mathcal{L}(j) = d_{\max},\forall j \in I$, where $\dist(\mathcal{L}(i); \mathcal{L}(i + k)) > 2\rIns$.  
  \item Instantaneous change in range:
    $|\mathcal{L}(i+1)-\mathcal{L}(i)| > 2\rIns$, 
\end{enumerate}
where $\rIns$ is the inscribed radius of robot and $\dist(\cdot; \cdot)$ is the distance in Cartesian space. Regions triggering the above are gaps that populate the gap
set $\gapSet$. Each gap in $\gapSet$ is then classified as swept or
radial, see Figure \ref{fig:GapGeom}.
Gaps picked based on the second test are automatically radial gaps. The
others are classified based on the dominant direction. 
Such direction is measured via the angle $\alpha$ on the short side of the
triangle formed by the robot and the two sides of the gap in the laser
scan, Figure \ref{fig:GapGeom}(d).  
Let $l_l = \mathcal{L}(i)$, $l_r = \mathcal{L}(i + k)$, and $l_{\min} =
\min[l_l, l_r]$. Then,
\begin{equation} \label{eq:skewnessangle}
  \small
    \alpha = \pi - 2\pi \frac{k}{n} 
    		 - \sin^{-1}{\frac{l_{\min}}{\sqrt{l_l^2 + l_r^2 - 2l_ll_r\cos(2\pi \frac{k}{n})} \sin(2\pi \frac{k}{n})}}.
\end{equation}
The threshold $\thAlpha = 3\pi/4$ differentiates radial from swept gaps.
Assign the gap type as {\em left} when $\angle l_r > \angle l_l$ and
{\em right} otherwise.
For each detected gap $\gap \in \gapSet$, record the angular index and
distance of its sides. 

\subsubsection{Gap Simplification}
A second pass through $\mathcal{G}$ merges radial gaps and replaces them
with swept gaps. This process is done by moving gaps into a stack
and recording the last continuously mergable radial gaps before reaching
a termination condition. The termination condition is defined by
parameter $c_{a}$ denoting the distance between the pending gap and
terminating gap. A gap is mergable with an incoming gap if
there are no closer obstacle points in between. 
Mergability is tuned by the parameter $c_{d}$, based on the acceptable range
difference between two merging candidate gaps.
By traversing every gap
once when added to the final set and possibly once more when determining
the need for deletion, the time complexity for removing redundant gaps
is linear.  We call this method {\em Swept Gap Prioritization} (SGP).

\begin{figure*}[t]
  \vspace*{0.06in}
    \centering
    \begin{tikzpicture}[inner sep=0pt, outer sep=0pt]
      \node[anchor=south west] (cg5) at (0in, 0in)
        {{\includegraphics[width=0.23\textwidth]{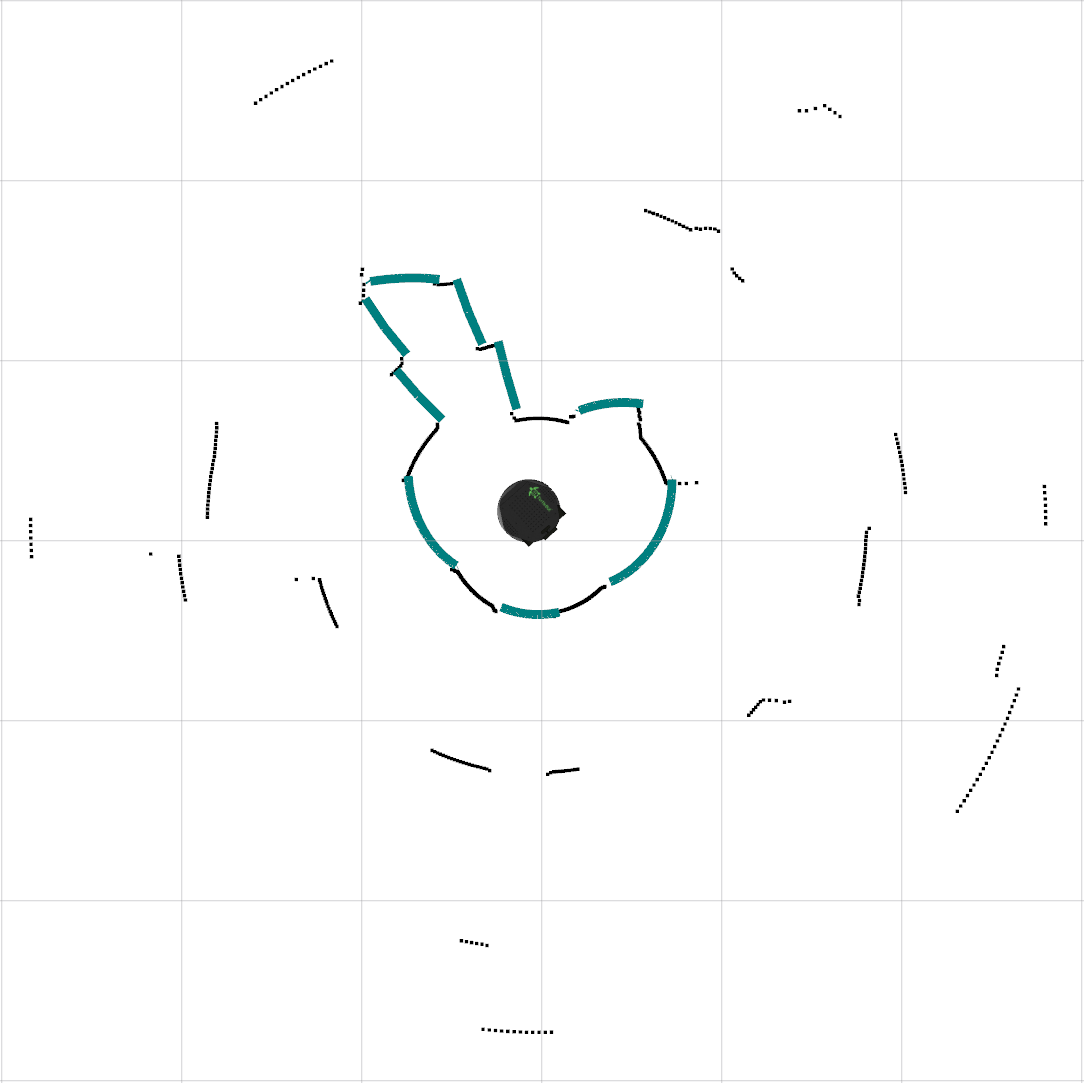}}};
      \node[anchor=south west] (rgp5) at (1.6in,0in)
        {{\includegraphics[width=0.23\textwidth]{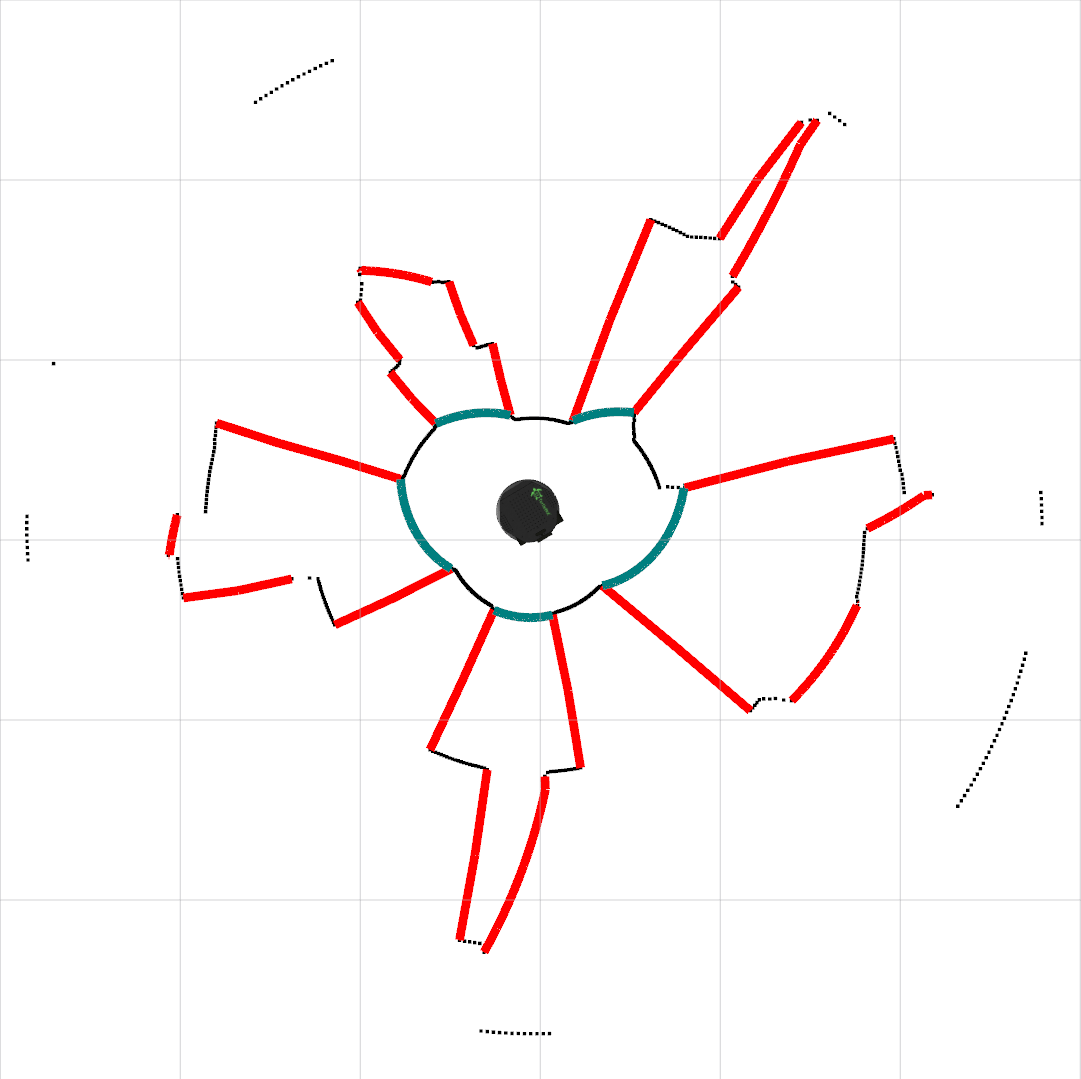}}};

      \node[anchor=south west] (cg8) at (3.2in,0in)
        {{\includegraphics[width=0.23\textwidth]{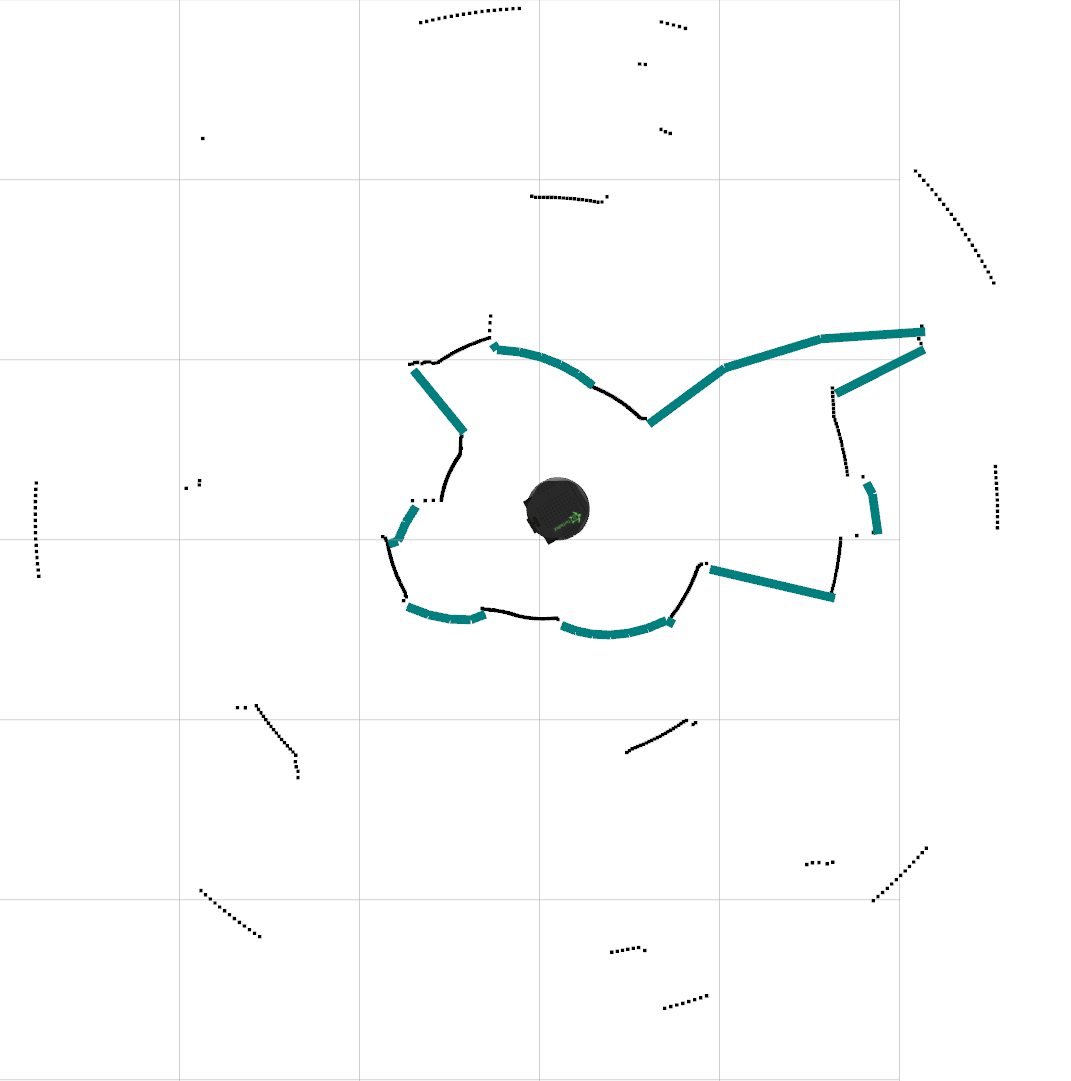}}};
      \node[anchor=south west] (rgp8) at (4.8in,0in)
        {{\includegraphics[width=0.23\textwidth]{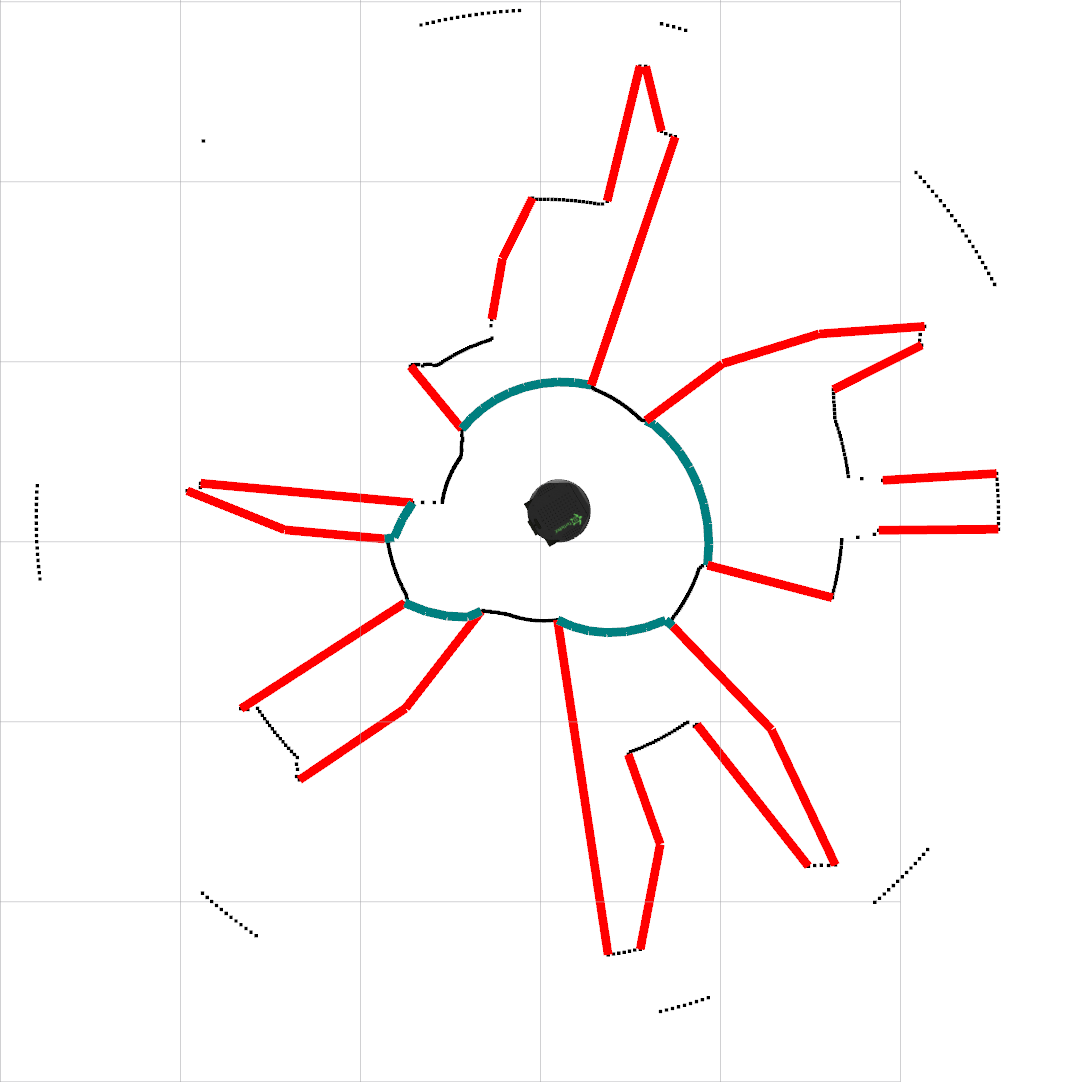}}};

      \node[anchor=north] at (cg5.north) {\footnotesize closest gap};
      \node[anchor=north] at (rgp5.north) {\footnotesize potential gap};
      \node[anchor=north] at (cg8.north) {\footnotesize closest gap};
      \node[anchor=north] at (rgp8.north) {\footnotesize potential gap};
      \node[anchor=south west] at (cg5.south west) 
        {\footnotesize (a) Closest gap vs potential gap.};
      \node[anchor=south west] at (cg8.south west) 
        {\footnotesize (b) Closest gap vs potential gap.};

      \node[anchor=south west] (f1cgD) 
        at ($(cg5.south west)+(1.2,3.2)$) {\textcolor{black}{D}};
      \node[anchor=south west] (f1rgpA) 
        at ($(rgp5.south west)+(3.2,3.6)$) {\textcolor{black}{A}};
      \node[anchor=south west] (f1rgpB) 
        at ($(rgp5.south west)+(3.5, 2.4)$) {\textcolor{black}{B}};
      \node[anchor=south west] (f1rgpC) 
        at ($(rgp5.south west)+(2,0.5)$) {\textcolor{black}{C}};
      \node[anchor=south west] (f1rgpD) 
        at ($(rgp5.south west)+(1.2,3.1)$) {\textcolor{black}{D}};

      \node[anchor=south west] (f2rgpE)  
        at ($(cg8.south west)+(3.4,3.0)$) {\textcolor{black}{E}};
      \node[anchor=south west] (f2cgE)  
        at ($(rgp8.south west)+(3.35, 2.95)$) {\textcolor{black}{E}};

      \node[anchor=north west,yshift=-5pt] (cg12) at (cg5.south west)
        {{\includegraphics[width=0.23\textwidth,clip=true,trim=0in 5in 0in 5in]{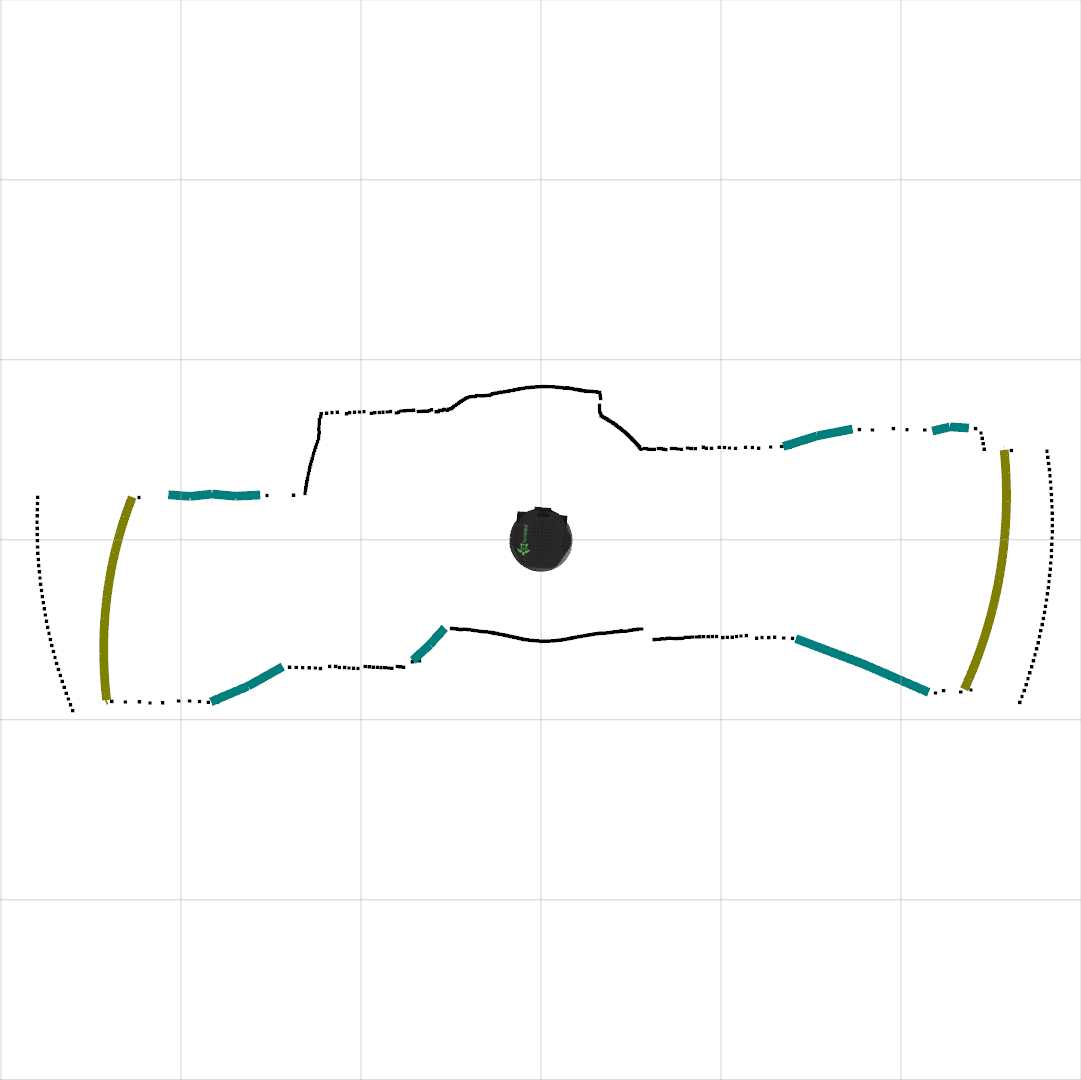}}};
      \node[anchor=north west,yshift=-5pt] (rgp12) at (rgp5.south west) 
        {{\includegraphics[width=0.23\textwidth,clip=true,trim=0in 5in 0in 5in]{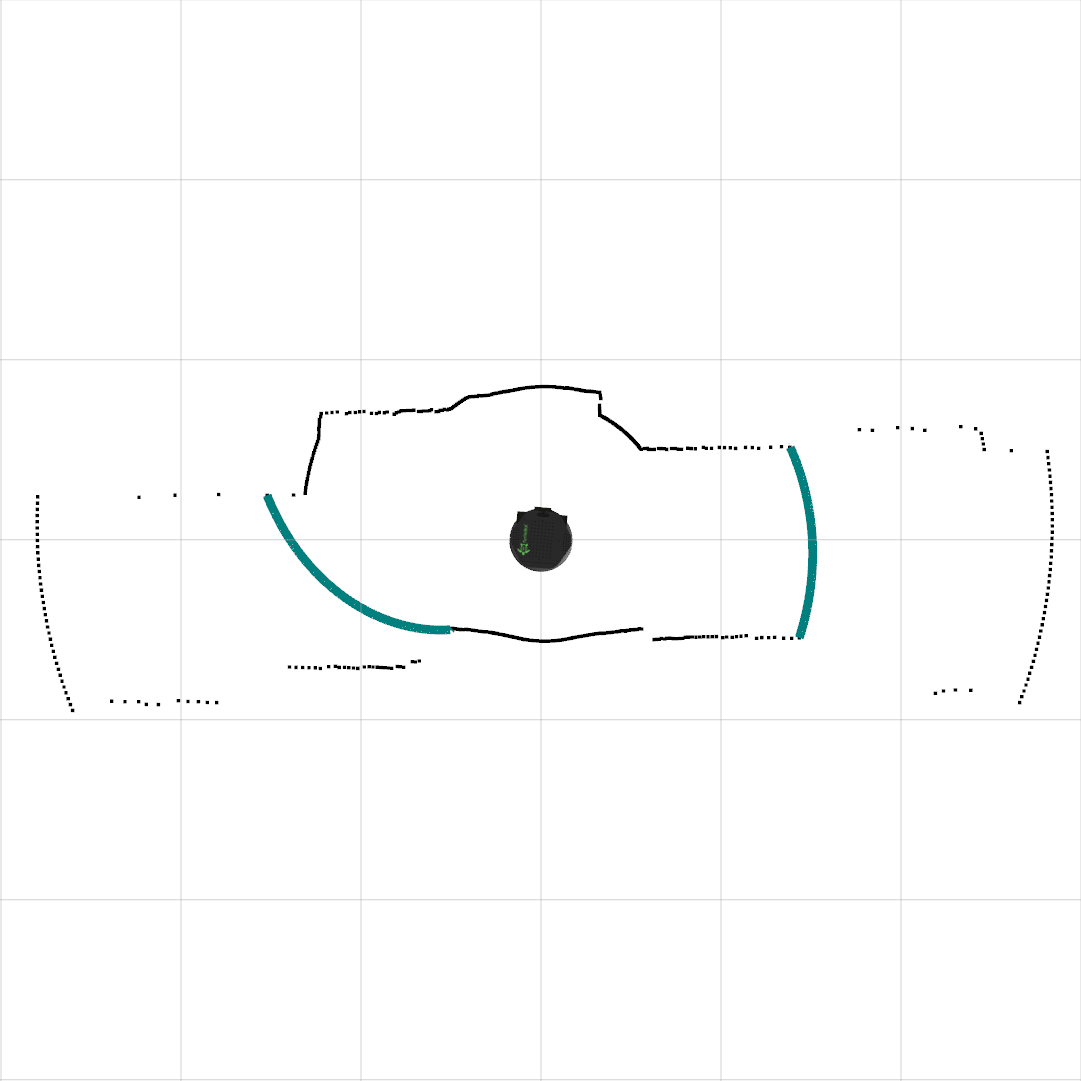}}};

      \node[anchor=north west,yshift=-5pt] (cg11) at (cg8.south west)
        {{\includegraphics[width=0.23\textwidth,clip=true,trim=0in 5.5in 0in 4.5in]
        {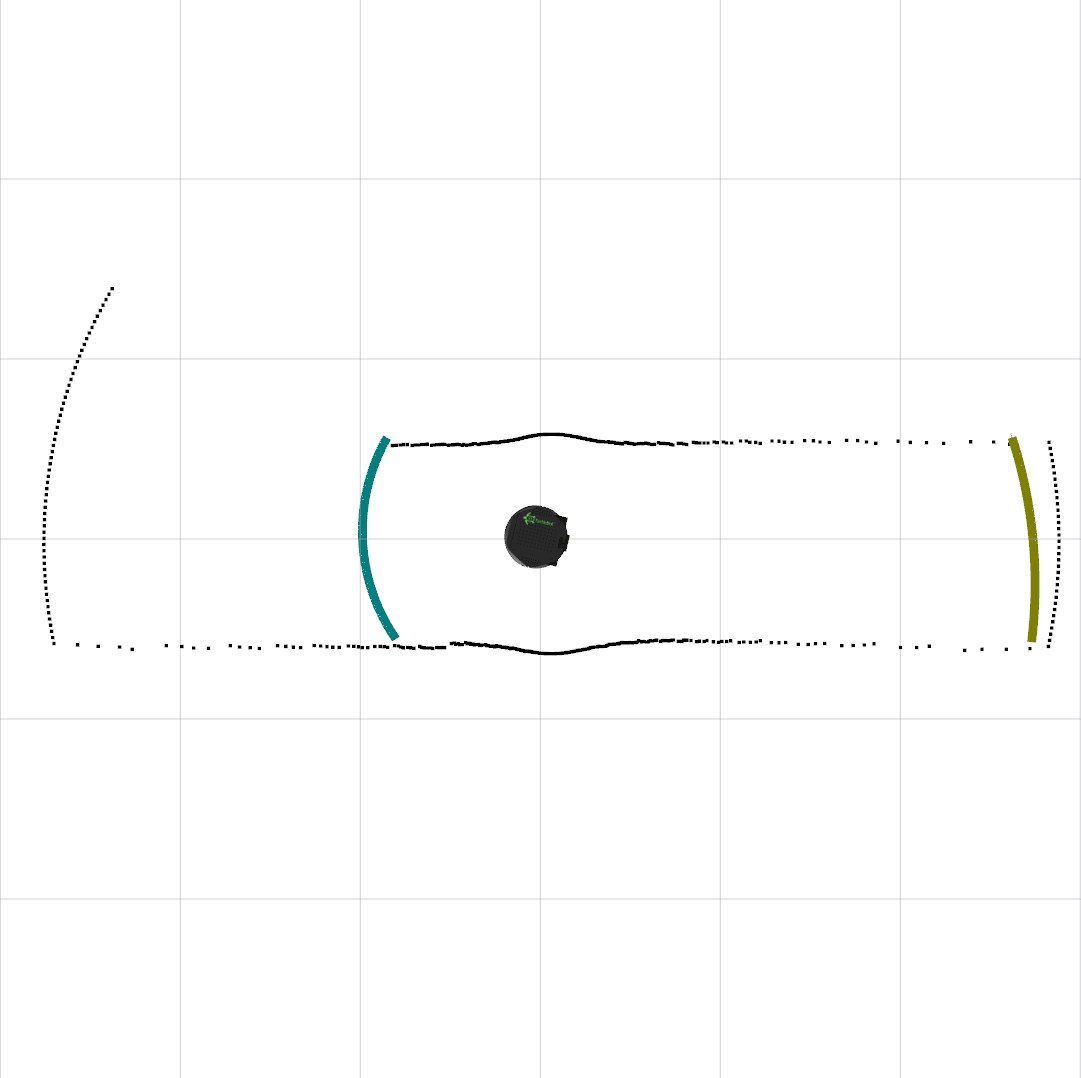}}};
      \node[anchor=north west,yshift=-5pt] (rgp11) at (rgp8.south west) 
        {{\includegraphics[width=0.23\textwidth,clip=true,trim=0in 5.5in 0in 4.5in]
        {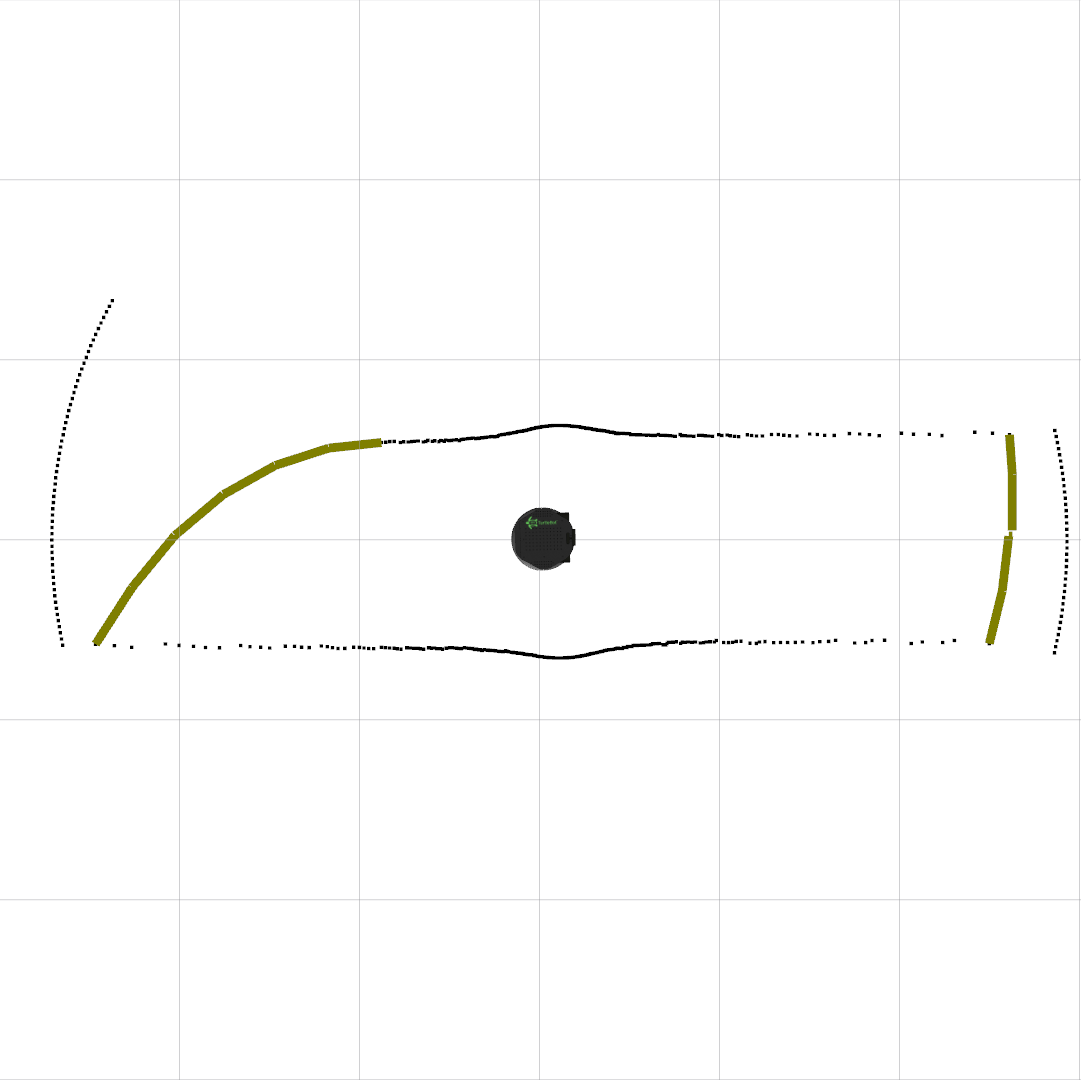}}};

      \node[anchor=south west] (f3cgF)  
        at ($(cg12.south west)+(0.8,1.0)$) {\textcolor{black}{F}};
      \node[anchor=south west] (f4rgpF) 
        at ($(rgp12.south west)+(0.8,1.0)$) {\textcolor{black}{F}};
      \node[anchor=north west] at (cg12.south west) 
        {\footnotesize (c) Closest gap vs potential gap.};
        
      \node[anchor=south west] (f3rgpG) 
        at ($(cg11.south west)+(1.1,1)$) {\textcolor{black}{G}};
      \node[anchor=south west] (f4cgF) 
        at ($(rgp11.south west)+(0.8,1)$) {\textcolor{black}{G}};
      \node[anchor=north west] at (cg11.south west) 
        {\footnotesize (d) Closest gap vs potential gap.};

    \end{tikzpicture}
    \caption{Comparison between closest gap (CG) \cite{CG} and potential
    gap (PG). 
    Dark dots represents laserscan results, red lines denote raw gaps
    for PG, and the teal lines denote returned gaps (CG and PG).
    {\bf (a)} Performing SGP collapses dangerous regions as A, B, and C. 
    {\bf (a,b)} Since PG explicitly reasons about swept gaps, it
    simplifies regions D and E while CG does not due to boundary
    geometry readings. 
    {\bf (c)} PG is less sensitive to sensor noise and provides a smaller, safer
    set of gaps as in region F, since regions behind swept gaps
    are collision free. 
    {\bf (d)} CG and PG are comparable for G.
    \label{fig:gap-parsing-comparison}}
\end{figure*}
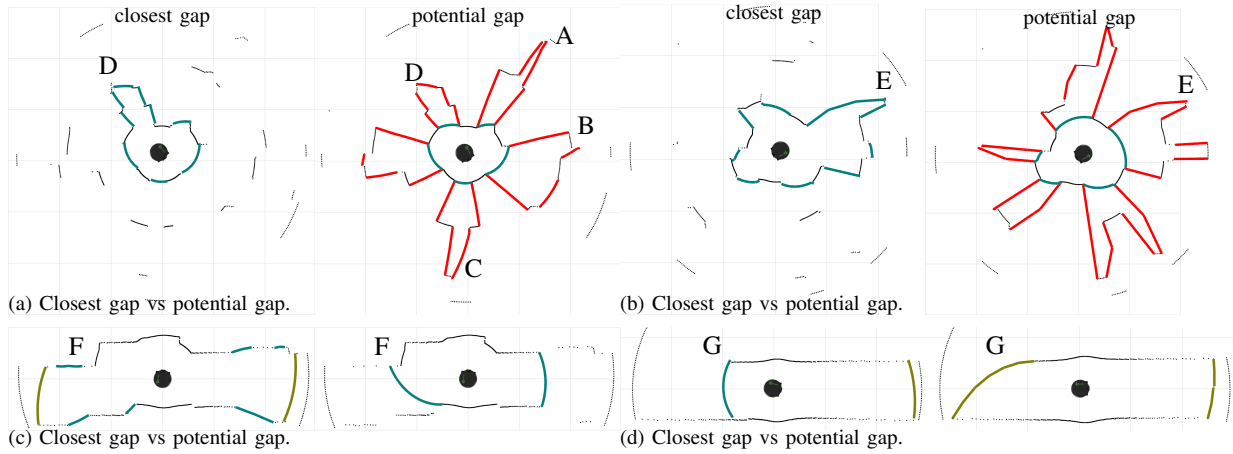

\subsubsection{Radial Gap Conversion}
For holonomic point-mass robots with full ($360^\circ$) scanning the SGP
strategy is sufficient for collision-free path planning.  However, for
robots with partial scanning, non-circular shapes, or nonholonomic
constraints, radial gaps cause problems based on the limited visual
information available about what lies beyond the gap. To remedy the
problem, a {\em radial gap conversion} process may be added to convert
all remaining radial gaps into swept gaps by rotating their gap
representation about the nearest gap point. The amount of rotation is
controlled by the parameters $\agP = (\agP^1,\agP^2)$ and by $\thAlpha$. 
Radial gaps are rotated by $\phi = \atan(\agP^2/\agP^1)$ so that the
local goal, a temporary point just $\agP^2$ away on the other side of
the gap is line-of-sight visible. 
Figure \ref{fig:GapGeom}(c) shows a preserved radial gap converted to a
swept gap. 

    



\subsubsection{Swept Gap Prioritization versus Other Gaps}
Figure \ref{fig:gap-parsing-comparison} visualizes some of the benefits
of potential gap (PG) over closest gap (CG) \cite{CG}, which attempts similar
processing.  The figures shows examples of gap detection where raw gaps
are marked bright red, and simplified gaps marked teal. PG collapses
compound situations (A, B, and C regions), reducing gaps to the ones
most relevant to the robot. The simplification reduces uncertainty
behind swept gaps and avoids dangerous compound scenarios such as at the
tip of region A and B.
While both approaches generate largely similar results in most scenarios,
PG is computationally simpler. It leverages gap positioning in polar
space during the reduction of gap sets and is more robust to situations
where sensors readings are uncertain.
Figure \ref{fig:gap-parsing-comparison}(c) illustrates where sensor
uncertainty due to environment geometry causes discontinuities in range
readings. PG simplifies some of the gaps and returns ones better suited
to motion planning.


\subsection{Gradient Field Construction} \label{sec:gapGradient}
The output of the RGP algorithm leads to a modified (and usually reduced)
gap set $\rgpSet$.
For each $\gap \in \rgpSet$ we establish a gap
local goal that is at least $\agP^2$ away from the gap,
and $\agP^1$ from the closest side of gap if the local goal
lies outside of the (visible) angular space of the gap.
The generation of a local navigation trajectory for
the gap involves the creation of attractive potentials whose gradients
are augmented with circulation terms \cite{DoMa[2003]GyroForce}. 
Under normal circumstances the addition of circulation would have sign
ambiguity.  Here, the desired flow direction to the gap is known, and hence
circulation sign, so that the constructed solution has guaranteed
passage through the gap.

\subsubsection{Closing the Gap for Convexity}
If the gap angular extent is beyond $\threshGA$ (usually $90$ or
$180^\circ$), then the gap angle is shrunk to give a new $\gapS \subset
\gap$ whose angular extent is $\threshGA$, thereby ensuring that the gap
region is a polar triangle convex in Euclidean space.  
Euclidean convexity ensures that the gradient vectors from the
potential and circulation elements will have the necessary properties to
guarantee gap passage. Shrinking of the gap is biased to ensure that the
local gap goal lies within the outer gap region.  The set $\rgpSet$ is
remapped to consist of Euclidean convex gaps only.


\subsubsection{Potential Field with Circulation}
After convexification, and possibly modification (see \S
\ref{secGapExtend}), to an angular extent equal to or less than
$\threshGA$, the local goal is line-of-sight visible to the other side
of the gap from any point within the gap region.  Let the local goal
point be $\xLG$ as determined from the chosen gap $G^* \in \rgpSet$.
The attractive potential is 
\begin{equation} \label{eq:dpot}
  \attPot(x) = \dist(x; \xLG) + \dist(x; \gap^*),
\end{equation}
where the first distance is to the local goal point 
and the second is to the gap curve (then vanishes on the other side of
the gap). These potentials attract the robot to the gap curve then
through to the local goal.  Rather than use the gradient for the flow,
we will follow the normalized gradient.

\begin{figure}[t]
  \vspace*{0.06in}
  \centering
  \begin{tikzpicture}[inner sep=0pt, outer sep=0pt]

    \node [anchor=south west] (circulation) at (0, 0)
    {{\includegraphics[width=0.3\columnwidth]{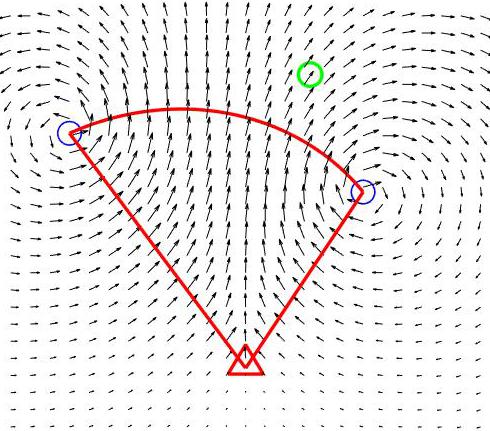}}};
    \node [anchor=south west,xshift=12pt] (attractor) at (circulation.south east)
    {{\includegraphics[width=0.3\columnwidth]{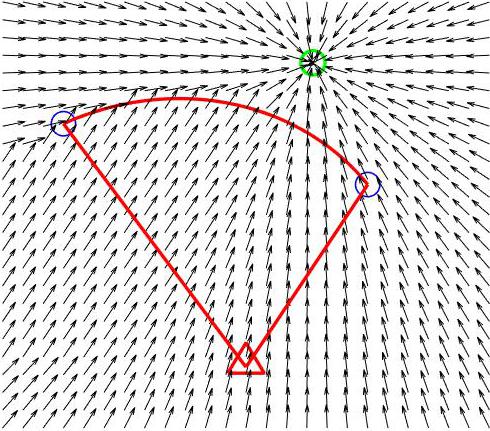}}};
    \node [anchor=south west,xshift=12pt] (combined) at (attractor.south east)
    {{\includegraphics[width=0.3\columnwidth]{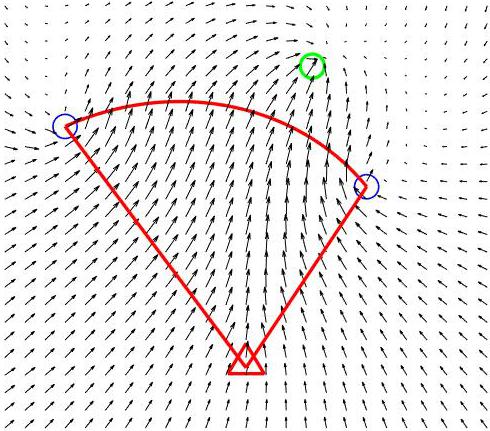}}};

    \node at ($(circulation.east)!0.5!(attractor.west)$) {+};
    \node at ($(attractor.east)!0.5!(combined.west)$) {=};
    \node[anchor=mid,yshift=0pt,fill=white] at (circulation.south) 
      {\scriptsize circulation};
    \node[anchor=mid,yshift=0pt,fill=white] at (attractor.south) 
      {\scriptsize attractor};
    \node[anchor=mid,yshift=0pt,fill=white] at (combined.south) 
      {\scriptsize combined};
  \end{tikzpicture}
  \caption{Gap gradient field construction. Red triangle is robot location, 
  blue circles are gap curve endpoints,
  and green circle is goal point.  \label{figGapGrad}}
\end{figure}
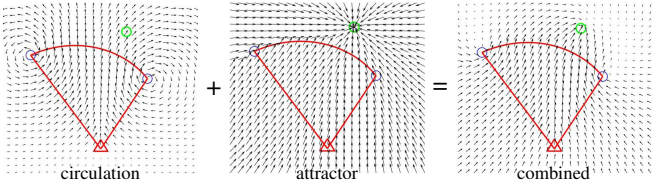

The premise behind gaps is that there is an obstacle in the world that
must be avoided by staying {\em within} the gap region, which is known
to be collision-free. Rather than impose an obstacle avoiding potential,
which could create a fixed point in the resulting vector field, a purely
rotational vector field is created 
\begin{equation} \label{eq:gcirc}
  \small
  \rotVF(x) = 
     \J e^{-\dist_\theta(x; \pLeft)/\bandGA}  
      \frac{\pLeft  - x}{||\pLeft  - x||} \\
     - \J e^{-\dist_\theta(x; \pRight)/\bandGA} 
       \frac{\pRight - x}{||\pRight - x||},
\end{equation}
where $\J = R(-\pi/2)$ is the $\mathfrak{so}(2)$ skew-symmetric form, and
$\dist_\theta(\cdot; \cdot)$ is the angular distance. The vector fields
are two rotational fields anchored at the left and right gap points.
Figure \ref{figGapGrad} shows an example circulation vector field. 

\subsubsection{Guaranteed Passage}
Proving that passage through the gap for the robot must happen involves
showing that the boundary of the gap navigation region points inwards
along the robot-to-gap-endpoint lines (or simply {\em gap lines}) and
that there is no fixed point interior to the region.  The only
reasonable flow for any point in the region is to exit via the gap
curve, e.g., {\em gap passage}. 
Along the gap lines, the gradients of \eqref{eq:dpot} either point
inwards or parallel to it, never out by virtue of gap convexity and by
definition of the gap region. Thus, what must be shown is that the
circulation components also point inwards. On the gap line, one
circulation term has $\dist_\theta$ vanishing; the circulation is purely
perpendicular and inward pointing. Let this vector be $e_\perp$. Let the
other circulation term contribute the vector $f_\phi$. It satisfies
\begin{equation}
  (e_\perp + f_\phi)\cdot e_\perp = (1 - \cos(\pi - \phi)) \ge 0,
  \ \text{for some}\ \phi > 0,
\end{equation}
which means that $\rotVF(\cdot)$ restricted to gap lines is inward
pointing.  The only outward flow can be on the gap curve.

A similar argument as above applies to show that the interior gap region
points have a non-trivial outward pointing flow. Define $e_\rho(x)$ to be
the radially directed outward vector for a point $x \in \gap$. The
following properties hold:
\begin{equation}
  e_\rho(x) \cdot \rotVF(x) \ge 0, \quad \text{and} \quad
  e_\rho(x) \cdot \hat \nabla \attPot(x) > 0
\end{equation}
by virtue of the gap angle being $90^\circ$ 
(see \S\ref{secGapExtend} for the $\threshGA = 180^\circ$ case and how
it gets further reduced to $90^\circ$). The definition of $\hat
\nabla$ is that it computes the gradient in \eqref{eq:dpot} then makes
it unit length.  The vector field interior to the gap region always has
positive outward pointing contributions, so there can be no fixed
points. Strict positivity implies that any initial point in $\gap$ will
flow out through the gap curve.



\newcommand{\traj}{\mathcal{T}}
\newcommand{\tpose}{p}
\newcommand{\tgoal}{p^*}
\newcommand{\egoCirc}{\mathcal{L}}
\subsection{Trajectory Synthesis and Scoring}
With the generated set of gaps from Section \S \ref{sec:gap-detect}, 
perform forward integration on every potential field to create a set of
trajectories $\mathcal{X}$. Each trajectory is scored through cumulative
egocentric pose cost and the terminating pose distance
to the local navigation goal
\cite{Smith2020}. Let $\traj$ denote a single trajectory
discretized into poses $\tpose$, $\tgoal$ denote the local navigation goal,
and $\dist(\tpose;\egoCirc)$ return the distance from $\tpose$ to the
nearest point in the egocircle $\egoCirc$. Score tuning parameters
include: $w_1, w_2, c_{\text{obs}}, r_{\text{max}}$. The trajectory score is:
\begin{equation}
  \begin{split}
    J(\traj) & = \sum_{\tpose \in \traj} 
        C(\dist(\tpose;\egoCirc)) + w_1||\tpose - \tgoal||\ , 
        \quad \text{where}\\
    C(d) & = \begin{cases}
      c_{\text{obs}} e^{-w_2 (d - \rIns)}, & d > \rIns \\
      0, & d > r_{\text{max}}\\
    \end{cases}
  \end{split}
\end{equation}
At each planning iteration, the best trajectory from current set $\mathcal{X}$
is selected and compared against the currently executing trajectory, with an
oscillation cost to limit path switching. 

\newcommand{\obSet}{\mathcal{C}}
\newcommand{\rCir}{r_{\text{cir}}}
\newcommand{\rMin}{r_{\text{min}}}
\newcommand{\rNom}{r_{\text{nom}}}
\subsection{Safe Extensions for Non-Ideal Robots \label{secGapSafety}}
The mathematics and algorithms employed to now presumed the use of an
idealized robot model. For deviations from this model, additional
modifications to the base method are required. This section describes
the modifications required for (a) robots with area/volume, (b)
noholonomic robots, and (c) other deviations from the ideal.

\subsubsection{Perception-Space Collision Checking}
This module is an automatic part of our perception-space navigation
stack \cite{Smith2017,Smith2020,SmEtAl[2020]egoTEB}. It takes
candidate local trajectories and synthesizes simulated {\em egocircle}
measurements at their future poses along the trajectory that measure
the far side of the robot. If any range measurements in the simulated
{\em egocircle} lie beyond the current actual {\em egocircle} estimate,
then the pose is in collision and the trajectory is rejected as
infeasible. The process repeats until a collision-free path is found
through a gap or all are rejected. The former gets passed on for
trajectory tracking while the latter triggers a global replan.

\subsubsection{Nonholonomic Vehicles}  The control $u \in \mathfrak{se}(2)$
is not fully realizable due to vehicle movement constraints. 
However, most nonholonomic vehicles (and certainly all standard mobile
robots) are linearly controllable along feasible trajectories.
Let the vehicle be a differential drive Hilare type.  The control is
converted to feasible movement via:
\begin{equation} \label{eq:NHctrl}
  \xi = [\nu, 0, \omega]^T , \ \nu = u^1, \ \omega = \lambda_y u^2 + u^3.
\end{equation}
where $\xi, u \in \mathfrak{se}(2)$.  The control is both realizable and
guides the robot towards the local gap goal. 

\subsubsection{Radial Extension of Gaps\label{secGapExtend}}
Since nonholonomy impacts the linear controllability of robot models, the
paths generated from the vector field in \S\ref{sec:gapGradient} cannot be
precisely followed, but are asymptotically tracked by \eqref{eq:NHctrl}.
The robot may actually leave the gap region when attempting to
do so (typically when oriented away or orthogonal to the local goal).
We address this problem by expanding the gap region in the vicinity of
the robot while closing it further from the robot. This process is
called {\em radial extension} since the gap region vertex is shifted
away from the gap curve along a radial line. Shifting the origin
permits the construction of a new gap region that puts the robot inside
of it. The robot may move for a non-zero distance and remain inside. 

\begin{figure}[t]
  \vspace*{0.06in}
  \centering
  \begin{tikzpicture}[inner sep=0pt, outer sep=0pt]
    \node[anchor=south west] (re) at (0in, 0in)
      {{\includegraphics[height=1in,clip=true,trim=1.25in 1.6in 3in 3.6in]{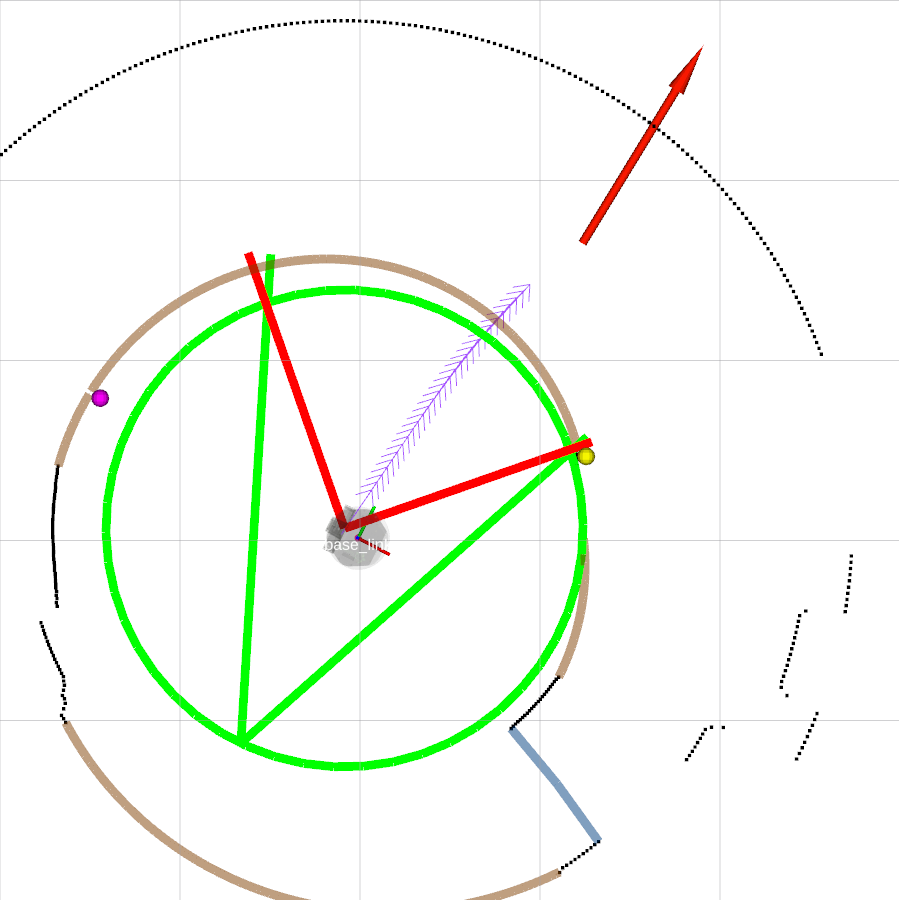}}};

    \node[anchor=north west] (osg) at (re.35) 
      {\footnotesize original polar triangle and swept gap (red)};
    \draw[->] ($(osg.west)-(2pt,0pt)$) -- ($(re.35)-(15pt,10pt)$);

    \node[anchor=west] (cfc) at (re.east) 
      {\footnotesize collision free circle (green)};
    \draw[->] ($(cfc.west)-(2pt,0pt)$) -- ($(re.east)-(12pt,0pt)$);

    \node[anchor=south west] (osg) at (re.330) 
      {\footnotesize new polar triangle and its swept gap (green)};
    \draw[->] ($(osg.west)-(2pt,0pt)$) -- ($(re)+(05pt,-11pt)$);

    \node[anchor=south west] (npo) at (re.south east) 
      {\footnotesize new polar origin};
    \draw[->] ($(npo.west)-(2pt,0pt)$) -- ($(npo.west)-(57pt,-2.5pt)$);
  \end{tikzpicture}
  \caption{Radial extension. The original swept gap polar triangle is
  converted to a new triangle with a shifted origin (backwards along the
  swept gap arc bisector). The shift distance is the radius of the
  largest free-space disc. \label{fig:radial-extension}}
\end{figure}
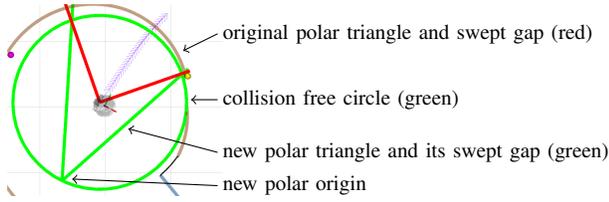

The process first finds the largest circular region centered on the robot
known to be collision-free.
Such a circle must exist if the robot is non-colliding.
The new origin is mapped to the intersection of the circle with the gap
boundary bisector that passes through the robot frame, see Figure
\ref{fig:radial-extension}. A new gap region is established from this new
origin and the radial vectors that pass through the two intersecting
points of the safe circle and the gap region boundary lines.
Simple geometric arguments show that the angular extent of this new gap is
half that of (or smaller than) the original gap. 
Gaps extending out to $\threshGA = 180$ are transformed to gaps with a
gap angle less than or equal to $90^\circ$, as preferred in \S\ref{sec:gapGradient}.


\subsubsection{The Safety Projection Operator} 
One final addition aims to prevent collisions, in case nonholonomic
motion and the discrete time implementation of continuous control laws
lead to collision for the edge cases. It implements the {\em projection
operator} from adaptive control, but to prevent entrance to a region as
opposed to preventing exit. It has long been used in adaptive control to
prevent parameter drift and is an early example of a barrier-like
function that performs command reshaping with provable forward
invariance for a convex set \cite{LaWi[2013]_RobAdaptCtrl}.  
Let $\obSet$ be the set of collision curves to the left and to the right
of the gap (in polar space) and $\dist(\cdot ; \obSet)$ the distance
function to $\obSet$.  Define the level-set function:
\begin{equation} \label{eq:poFun}
  \psi(x) 
    = \left(\frac{\rMin}{\dist(x; \obSet)} - \frac{\rMin}{\rNom} \right)/
      \left( 1 - \frac{\rMin}{\rNom}\right),\ 
\end{equation}
where $\rMin < \rNom$ and which is depicted in Figure \ref{fObPot}. 
Far enough away, the potential is
negative. At $\rNom$, $\psi$ passes through zero and at $\rMin$ it passes
through unity. These properties are essential to the projection operator,
which is defined as in \eqref{eq:ProjOp}. In adaptive control, the
projection operator keeps adaptive parameters from going unbounded
by restricting the parameters to a compact set. Here, by definition of
$\psi$, the opposite holds with the intent being to drive the state
away from the open, bounded set associated to $\Omega_O = \left\{ x |
\psi(x) > 1 \right\}$. When $\psi$ evaluates to less than zero, no
control modification occurs. When greater than zero, but with the control
vectored away from the boundary, no control modification occurs. As
$\psi \rightarrow 1$ under a violating law $u$, the control
modification increases until canceling out the component leading to
the safety violation. Under full control, the projection operator
prevents control dynamics that would enter the unsafe set $\Omega_O$
established by the obstacles.

\begin{figure*}[t]
  \begin{minipage}{0.75\textwidth}
  \begin{equation} \label{eq:ProjOp}
    \proj(u; x) = \begin{cases}
       u & \text{if}\ \psi(x) < 0 \\
       u & \text{if}\ \psi(x) \ge 0 \ \wedge\  
         \left< \frac{\nabla \psi(x)}{\left\| \nabla \psi(x) \right\|}, u \right> 
         > 0 \\
       u - \psi(x) \left< \frac{\nabla \psi(x)}{\left\| \nabla \psi(x)
       \right\|}, u \right> \frac{\nabla \psi(x)}{\left\| \nabla \psi(x)
            \right\|}
       & \text{if}\ \psi(x) \ge 0 \ \wedge\  
         \left< \frac{\nabla \psi(x)}{\left\| \nabla \psi(x) \right\|}, u \right> 
         \le 0 \\
       \end{cases}
  \end{equation}
  \end{minipage}
  \hfill
  \begin{minipage}{0.210\textwidth}
    \vspace*{0.06in}
    \centering
    \begin{tikzpicture}[inner sep=0pt,outer sep=0pt]
	  \node[anchor=south west] (pop) at (0in, 0in)
      {{\includegraphics[height=0.75in,clip=true,trim=0.25in 0.05in
      0.25in 0.25in]{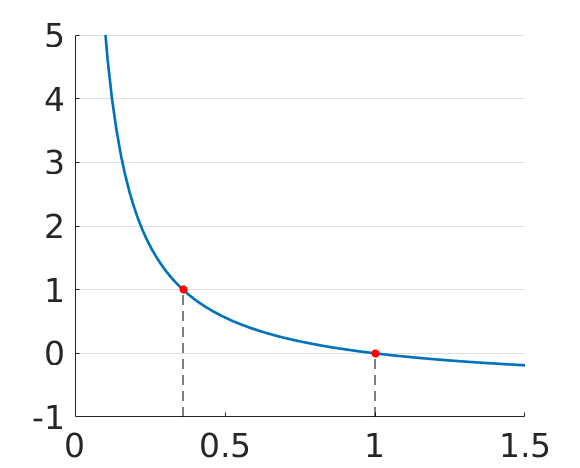}}};
      
      \node[anchor=north west] (rmin) at ($(pop.south west) + (17pt,30pt)$) 
        {\footnotesize $\rMin$};
      \node[anchor=north west] (rnorm) at ($(pop.south west) + (40pt,21pt)$) 
        {\footnotesize $\rNom$};
      
      \node[anchor=south west,xshift=2pt,yshift=-1pt] (xlab) at (pop.south east)
        {\footnotesize $\dist(x; \obSet)$};
      \node[anchor=east,xshift=-1pt] (ylab) at (pop.west) {\footnotesize $\psi$};
    \end{tikzpicture}
    \vspace*{-7pt}
    \caption{Obstacle Potential \label{fObPot}}
  \end{minipage}
  \vspace*{-0.5em}
\end{figure*}
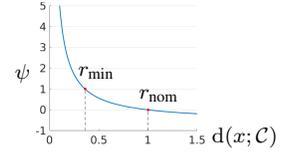

\section{Experiments, Outcomes, and Discussion}
This section covers the experiments and outcomes conducted for potential
gap and for the state-of-the-art method TEB. Monte Carlo rollouts are
performed for navigation tests in benchmark worlds, which are populated
with randomized unknown obstacles and start/goal locations. 

\subsection{Benchmark Environments}
The TEB and {\em potential gap} (PG) navigation planner are run in four
benchmarking worlds. 
Dense, Campus, and Office are from \cite{SmEtAl[2020]egoTEB}. 
The fourth, Sector world, is a single large room intended to model
indoor spaces with locally untraversable regions such as chair legs and
obstacles such as trash cans and cabinets. 
The simulation environment is a slightly modified version of the Simple Two
Dimensional Robot Simulator (STDR)
\cite{stdr_simulator}. A very short range
360$^\circ$ range finder acts as a virtual bumper for collision detection. 
The platforms are configurable with different angular fields of view (FoV) 
of the laser sensor, starting from 60$^\circ$ to 360$^\circ$, 
for partial to full observability. Additionally, experiments are
conducted in Gazebo with Turtlebots (60$^\circ$ FoV) for navigation
performance with nonholonomic, 2$^{\rm nd}$ order robot.



\subsection{Experiments and Outcomes}
The experiments cover a range of test cases, starting with the idealized
robot model under variable field of view (FoV) settings. The idealized
robot model has 1$^{\rm st}$ order, fully actuated control equations
(i.e., holonomic) with a circular body that operates like a
point mass. The second experiment relaxes the control equations to being
nonholonomic. The third experiment moves from STDR to Gazebo, where the
motion model is nonholonomic, 2$^{\rm nd}$ order.
Each setting is run with 100 different seeds in all four worlds.
STDR experiments are run 25 at a time and Gazebo experiments
3 at a time on an AMD Ryzen 7 3800X processor (single-thread passmark
of 2744; multi-thread score of 43,447). 


\begin{figure*}[t]
  \begin{minipage}[t]{1.47\columnwidth}
    \begin{tikzpicture}
      \node[anchor=west] (pg) {
        \includegraphics[width=0.32\columnwidth]{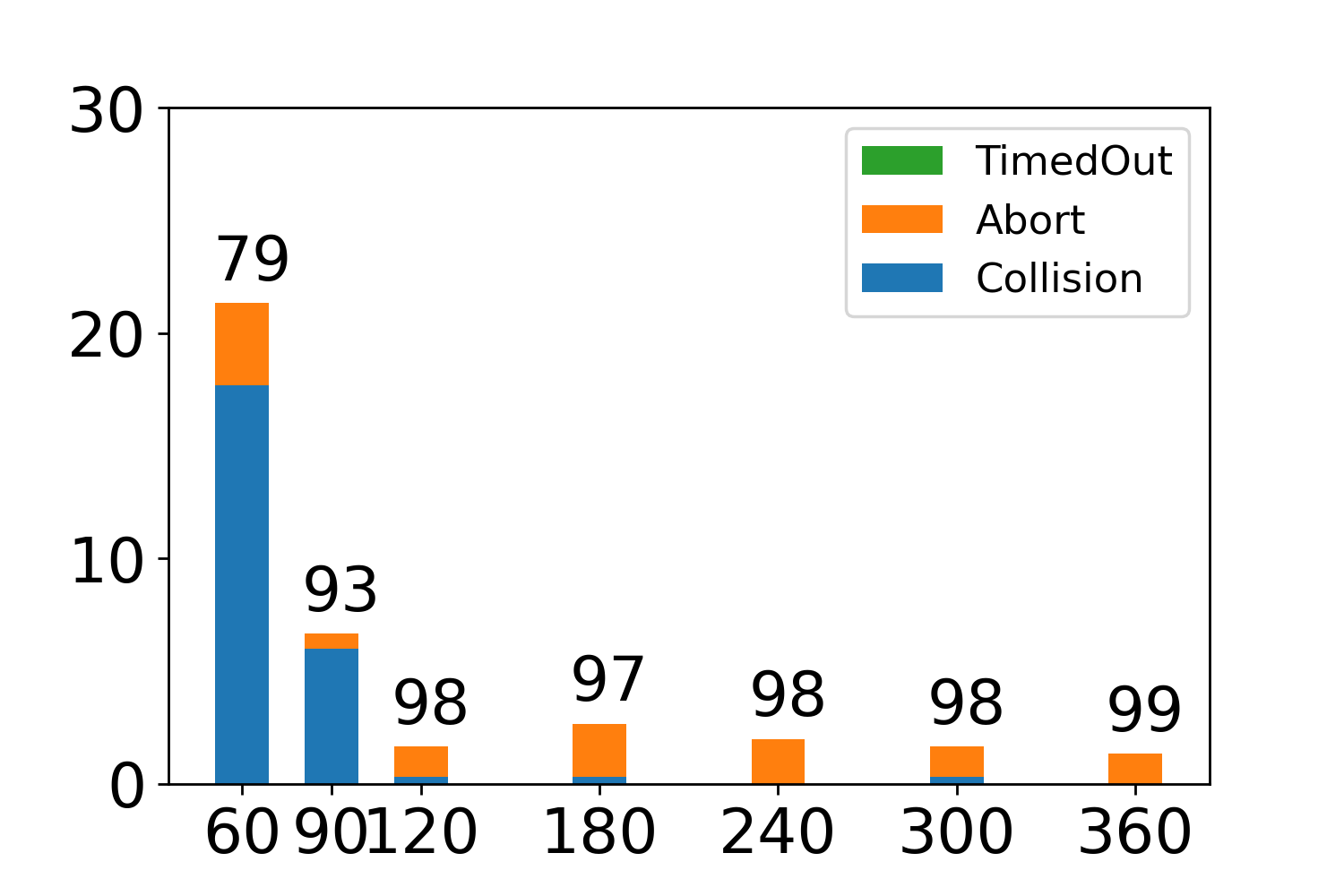}
        };
      \node[anchor=west] (pg_pore) at (pg.east) {
        \includegraphics[width=0.32\columnwidth]{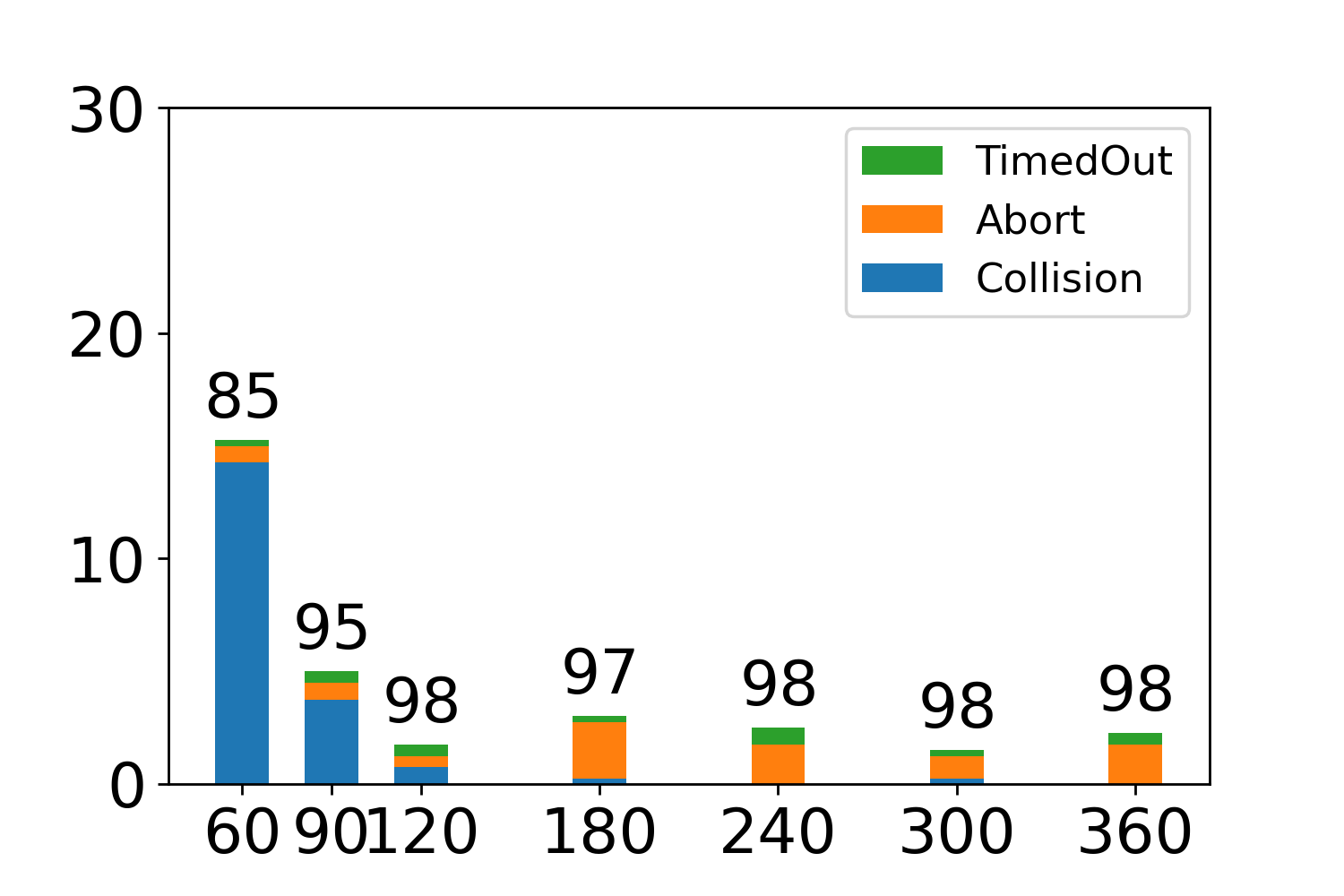}
      };
      \node[anchor=north west] (pg_title) at ($(pg.north west)+(18pt,-11pt)$) 
        {\scriptsize PG};
      \node[anchor=north] (pg_xaxis) at ($(pg.south) + (0, 0.2)$) {\footnotesize FoV};
      \node[anchor=south, rotate=90] (pg_yaxis) at ($(pg.west) + (0.3, 0)$) {\footnotesize Percentage};

      \node[anchor=west] (teb) at (pg_pore.east) {
        \includegraphics[width=0.32\columnwidth]{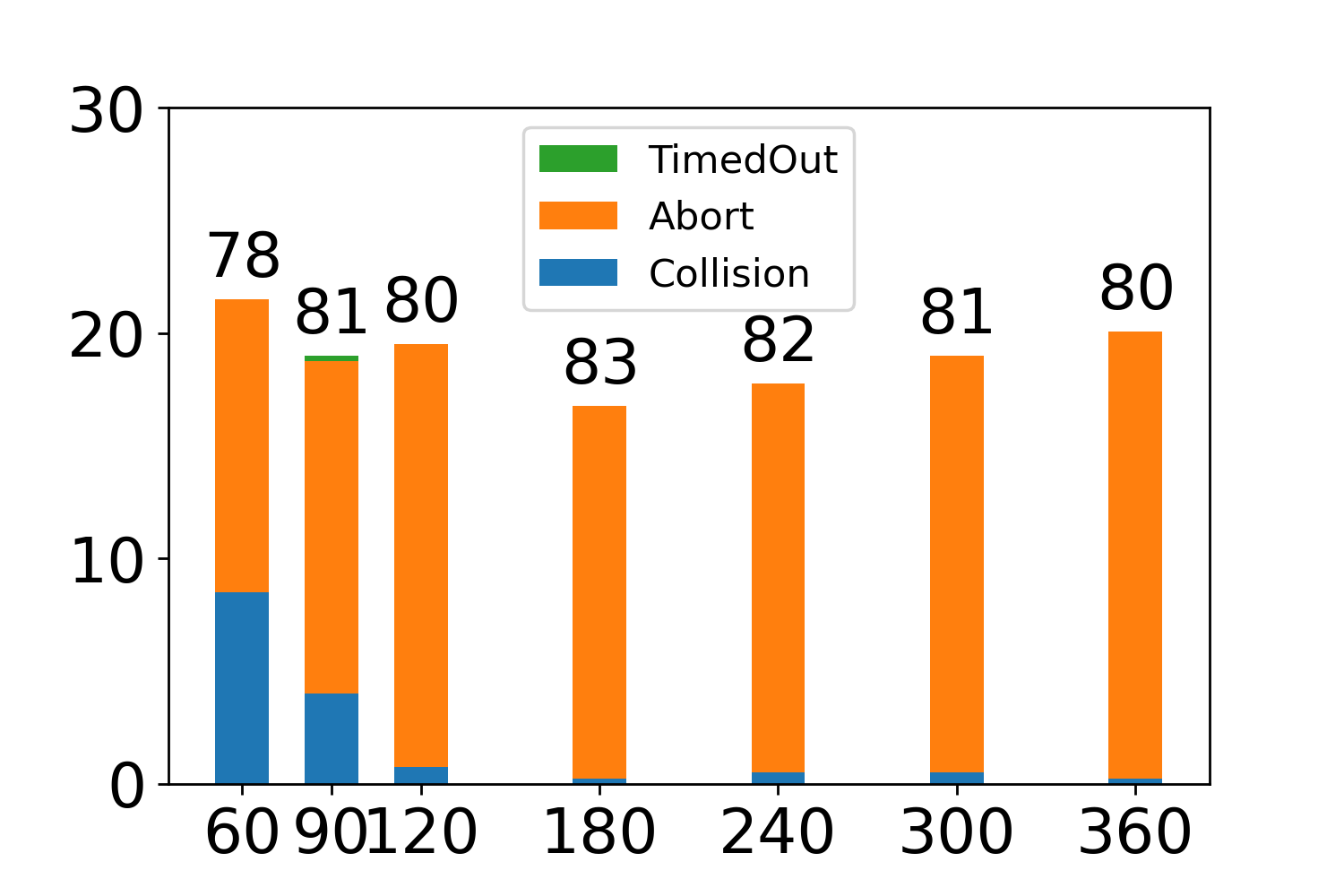}
      };
      \node[anchor=north west] (teb_title) at ($(teb.north west)+(18pt,-11pt)$) 
        {\scriptsize TEB};
      \node[anchor=north] (teb_xaxis) at ($(teb.south) + (0, 0.2)$) {\footnotesize FoV};
      \node[anchor=south, rotate=90] (teb_yaxis) at ($(teb.west) + (0.3, 0)$) {\footnotesize Percentage};

      \node[anchor=north west] (pg_pr_title) at ($(pg_pore.north west)+(18pt,-11pt)$) 
        {\scriptsize PG+RE};
      \node[anchor=north] (pg_pr_xaxis) at ($(pg_pore.south) + (0, 0.2)$) {\footnotesize FoV};
      \node[anchor=south, rotate=90] (pg_pr_yaxis) at ($(pg_pore.west) + (0.3, 0)$) {\footnotesize Percentage};
    \end{tikzpicture}
    \caption{PG (left) and TEB (right) performance vs FoV for the
    idealized robot model ($1^{\rm st}$-order holonomic). Numbers over
    columns are success rate (\%). PG+RE (middle) adds radial extension
    to address FoV issues.
    \label{fig:holoRes}}
  \end{minipage}
  \hfill
  \begin{minipage}[t]{0.52\columnwidth}
  \begin{tikzpicture}
    \node[anchor=north west] (picture) {
        \includegraphics[width=0.9\columnwidth,clip=true,trim=0.65in 45 0 45]{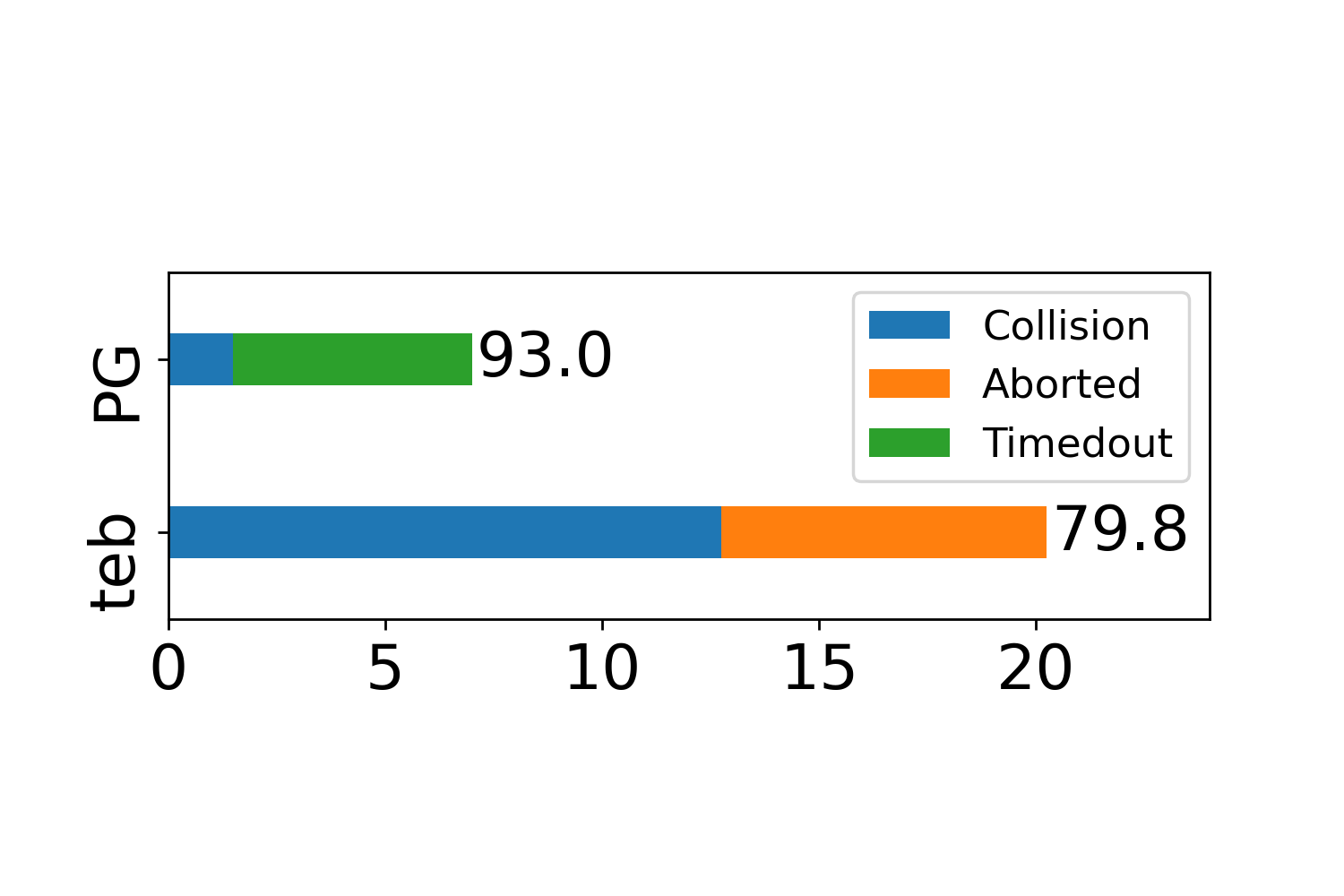}
      };
    \node[anchor=north,xshift=-1em,yshift=8pt] (xaxis) at ($(picture.south)$) {\footnotesize Percentage};
    \node[anchor=east,xshift=7pt,yshift=8.5pt] at (picture.west) {\footnotesize PG};
    \node[anchor=east,xshift=7pt,yshift=-9pt] at (picture.west) {\footnotesize TEB};
  \end{tikzpicture}%
  \addtocounter{figure}{1}%
  \caption{PG and TEB performance for fixed FoV in Gazebo.
  \label{fig:gazeboRes}}
  \end{minipage}
\end{figure*}

\subsubsection{Idealized Robot Model\label{sss:holo}}
Figure \ref{fig:holoRes} gives the failure mode vs FoV for PG and TEB
implementations. Firstly, for the full FoV case, PG achieves
no collisions and a nearly perfect success rate.
A small quantity (1\%) of runs resulted in an abort, which means
that the local and global planner could not coordinate to find a
feasible path to the goal. PG continues to match or
outperform TEB in collisions, aborts, and time outs in lower FoV cases. PG also maintains a
high success rate, much like TEB preserves it moderate success rate.
However, for 60$^\circ$ and 90$^\circ$ FoV, PG has more collisions,
and matches TEB's success rate at the lowest FoV. Enabling radial
extension gives a bit more navigation free-space around the robot to
compensate for the limited FoV.  PG+RE matches or outperforms TEB for
all cases except for 60$^\circ$, where the collision count remains higher.

The results show that in the idealized setting, the PG navigation system
matches the theoretical expectations. While TEB also has similar
performance, PG employs a greedy gradient descent approach to path
generation. In contrast, TEB employs a soft-constraint optimal control
formulation to establish the local navigation paths and has higher
computational cost.
The lower FoV results involve collisions that arose from a lack of 
FoV but can be corrected with the RE module. The high success rate of PG
and PG+RE confirms that the notion of using gaps to serve as navigation
affordances is critical to achieving safe, goal attaining navigation
through an unknown environment. 

\begin{figure}[t]
  \begin{tikzpicture}
    \node(coll) {
      \includegraphics[width=0.49\columnwidth]{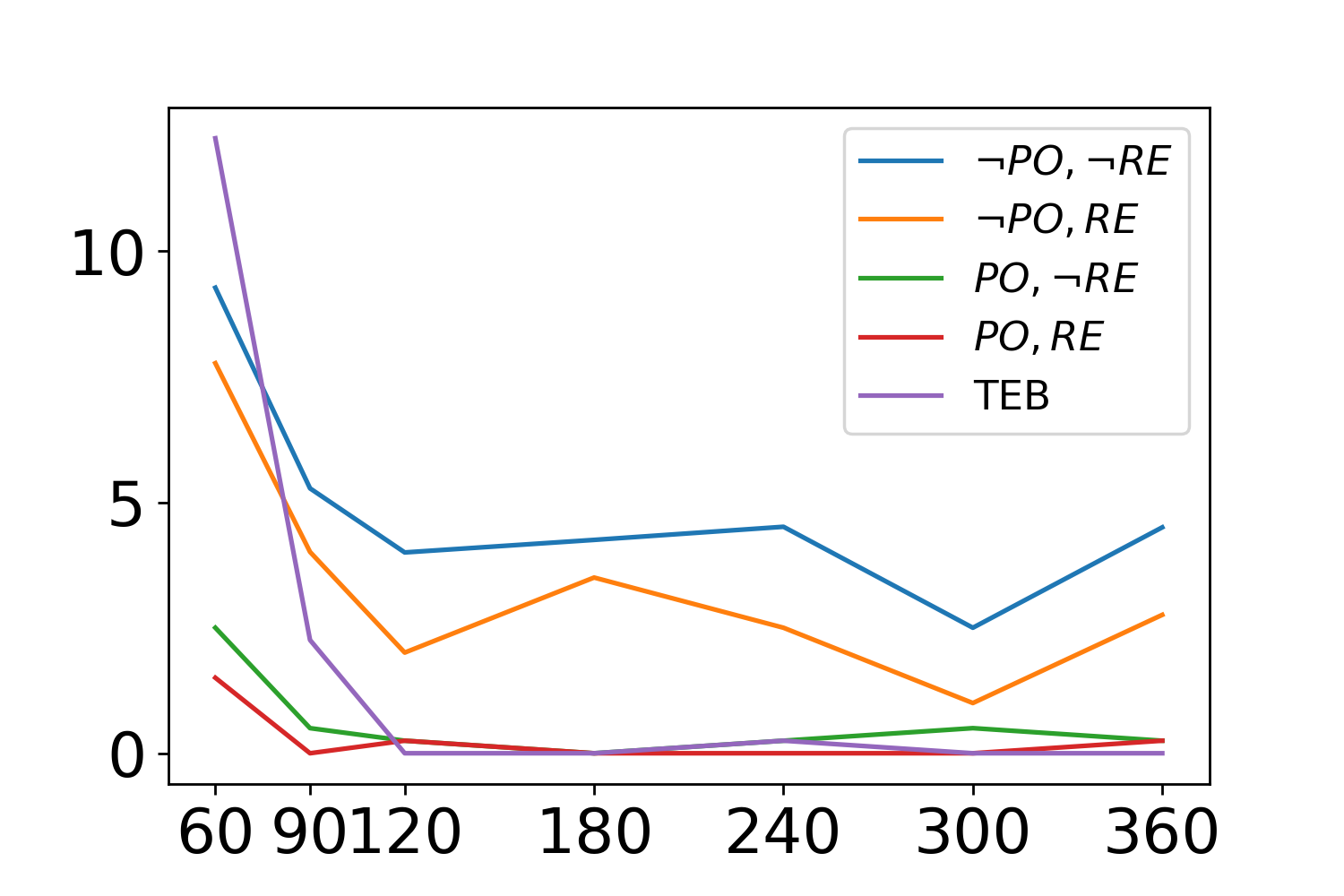}
    };
    \node[anchor=west] (succ) at (coll.east) {
      \includegraphics[width=0.49\columnwidth]{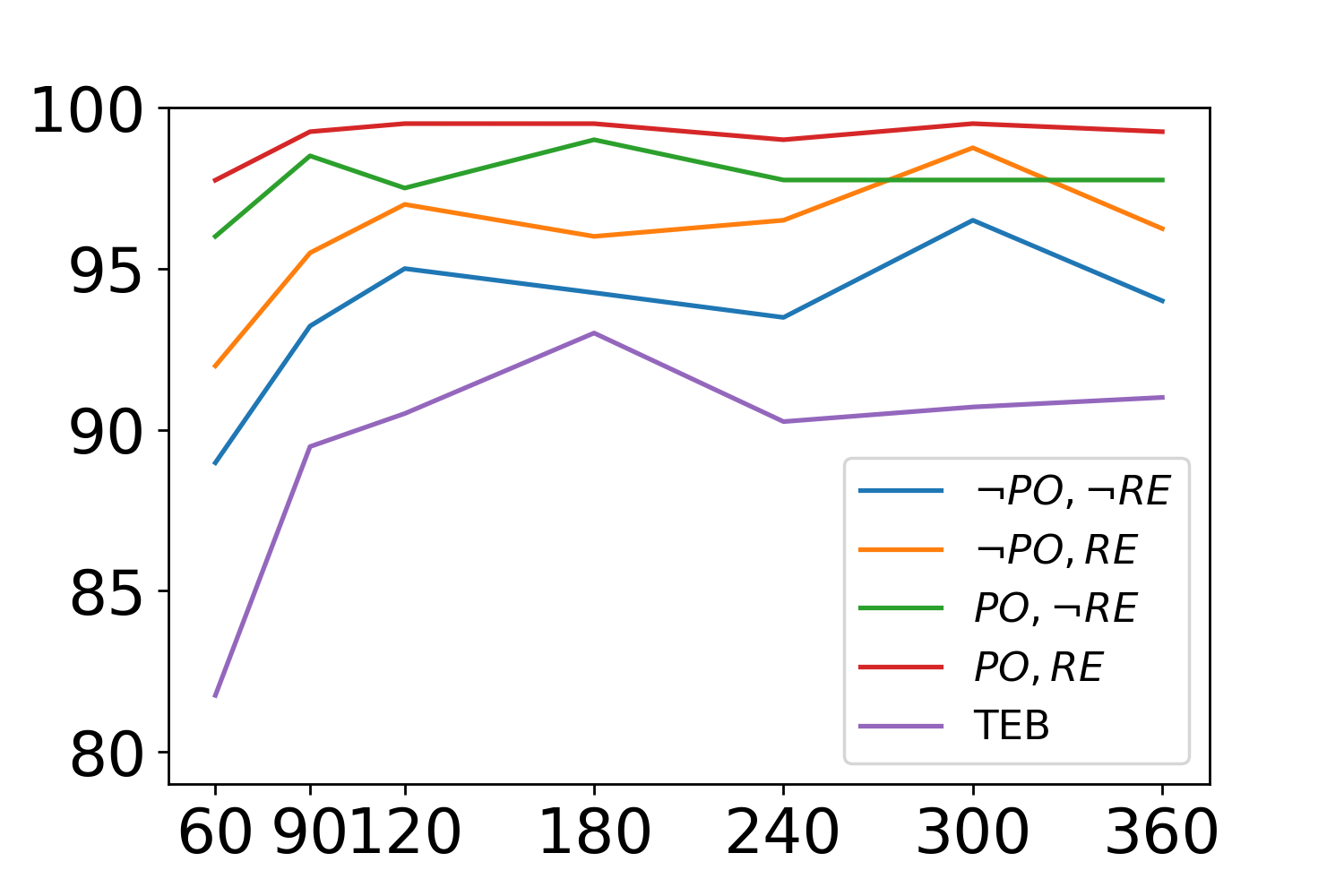}
    };
    \node[anchor=south] (coll_title) at ($(coll.north) + (0, -0.5)$) {\footnotesize Bumper Collisions};
    \node[anchor=north] (coll_xaxis) at ($(coll.south) + (0, 0.2)$) {\footnotesize FoV};
    \node[anchor=south, rotate=90] (coll_yaxis) at ($(coll.west) + (0.3, 0)$) {\footnotesize Percentage};
    \node[anchor=south] (succ_title) at ($(succ.north) + (0, -0.5)$) {\footnotesize Success Rate};
    \node[anchor=north] (succ_xaxis) at ($(succ.south) + (0, 0.2)$) {\footnotesize FoV};
    \node[anchor=south,rotate=90] (succ_yaxis) at ($(succ.west) + (0.3, 0)$) {\footnotesize Percentage};
  \end{tikzpicture}
  \addtocounter{figure}{-2}%
  \caption{PG and TEB bumper collisions (left) and success rate (right) vs FOV 
  for the $1^{\rm st}$ order nonholonomic case.
  \label{fig:nonholoRes}}
  \addtocounter{figure}{1}%
\end{figure}

\subsubsection{Nonholonomic, 1$^{\rm st}$ order\label{sss:nonholo1}}
Figure \ref{fig:nonholoRes} provides the outcomes for bumper
collisions (left) and success rate (right). Each plot contains
implementations of TEB, baseline PG without projection operator ($\neg$PO) and
radial extend ($\neg$RE), the full nonholonomic (PO,RE),
and mixed activations. The baseline PG, which uses the
idealized robot solutions, has the worst performance in lower FoVs.
TEB maintains a low collision count and has none at
360$^\circ$ FoV. Of the two partially corrected PG implementations,
the projection operator performs best (PO,$\neg$RE). Adding radial
extension improves it somewhat for higher FoV, showing that both
contribute to safe navigation. The zero collision performance of TEB at
full FoV is undermined by its low success rate induced by aborts.
Furthermore, the PG+(PO,RE) method has more consistent success rate as a
function of FoV.

Similar collision outcomes between PG+(PO,RE) and TEB 
show that the modified method functions as intended and that the
greedily derived paths are similar to optimally derived paths from a
safety perspective. They are better than the sampled and optimally
derived paths of TEB from a task completion perspective.
Compared to the holonomic case in Figure \ref{fig:holoRes}, 
both planners displayed an outcome less sensitive to FoV.
This is caused by the motion model limiting motion to angles local to
the camera optical axis.


\subsubsection{Nonholonomic, 2$^{\rm nd}$ order}
Figure \ref{fig:gazeboRes} displays the failure rates for PG+(PO,RE) and
TEB in the Gazebo simulated Turtlebot (FoV = 60$^\circ$). The TEB
collision rate matches the 1$^{\rm st}$ order case. The 2$^{\rm nd}$
order nonholonomic PG collision rate is similar to the 1$^{\rm st}$
order case and significantly better than TEB ($\sim$6x).  The success
rate for PG drops from 96\% to 93\%, but is well above those of 
the 1$^{\rm st}$ and 2$^{\rm nd}$ order TEB cases.

PG is relatively consistent when applied to the dynamic case.
Again, this is an important outcome based on the fact that PG is based
on APF methods that were designed for idealized robot models. As a
greedy path generation approach, it outperforms the soft-constraint
optimal control TEB solution. Though TEB explores a far richer
trajectory-space for a given local goal, the extra degrees of freedom
may not properly exploit the navigable space as well as the potential
gap method does. Especially since PG typically explores 5x more
trajectory options than TEB.
Furthermore, the soft constraint approach cannot guarantee feasibility
of the trajectories, which may result in poor sequential decision making
if some of the solutions are immediately feasible but eventually
infeasible. By construction PG avoids these issues and creates
protective mechanisms to prevent collision while aiming to complete the
task.

\begin{figure}[t]
  \begin{tikzpicture}
    \node(part) {
      \includegraphics[width=0.49\columnwidth]{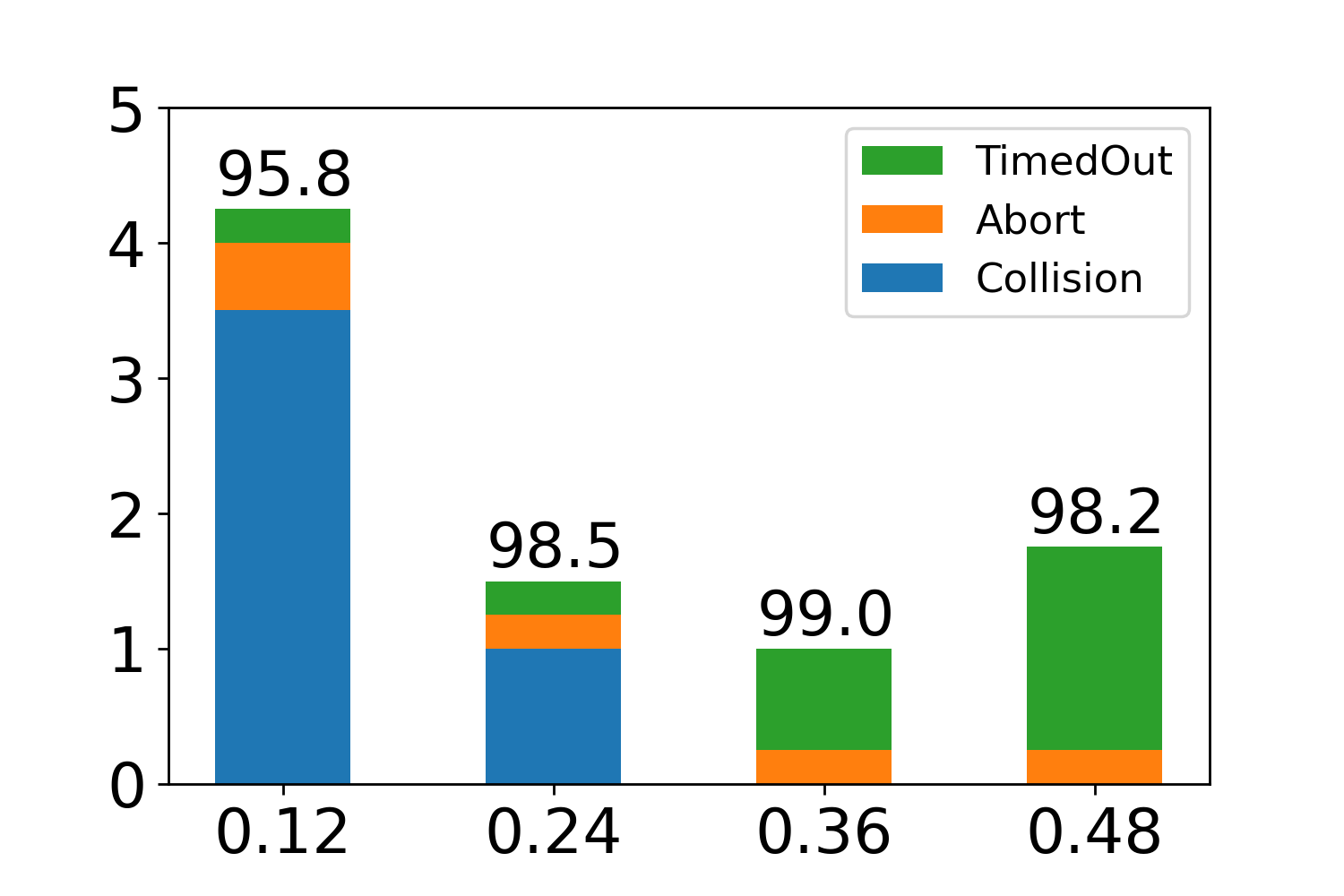}
    };
    \node[anchor=west] (full) at (part.east) {
      \includegraphics[width=0.49\columnwidth]{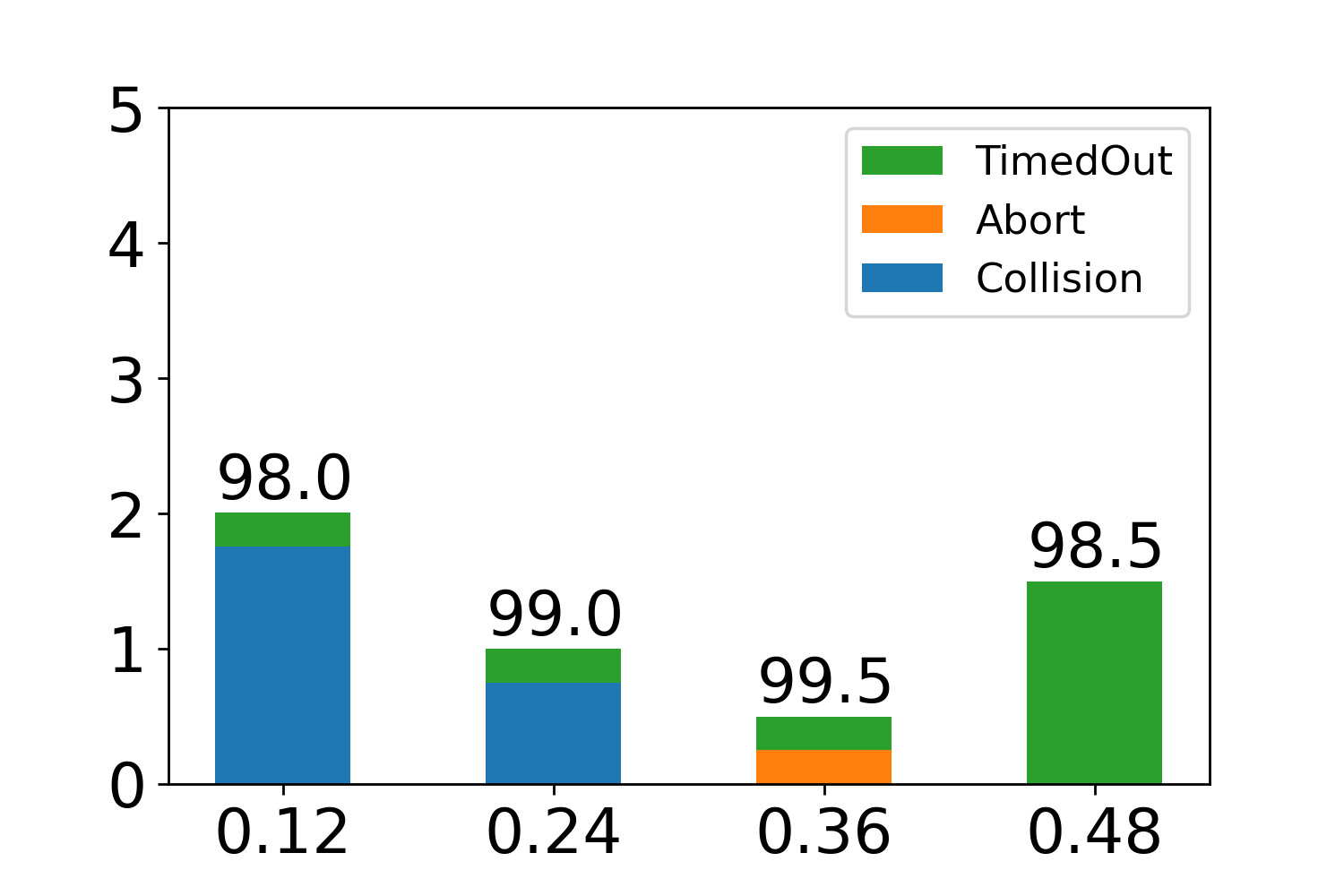}
    };
    \node[anchor=south] (part_title) at ($(part.north) + (0, -0.5)$) {\footnotesize $\rMin$ on Partial FoV};
    \node[anchor=north] (part_xaxis) at ($(part.south) + (0, 0.2)$)
    {\footnotesize $\rMin$};
    \node[anchor=south, rotate=90] (part_yaxis) at ($(part.west) + (0.3, 0)$) {\footnotesize Percentage};
    \node[anchor=south] (full_title) at ($(full.north) + (0, -0.5)$) {\footnotesize $\rMin$ on Full FoV};
    \node[anchor=north] (full_xaxis) at ($(full.south) + (0, 0.2)$)
    {\footnotesize $\rMin$};
    \node[anchor=south, rotate=90] (full_yaxis) at ($(full.west) + (0.3, 0)$) {\footnotesize Percentage};
  \end{tikzpicture}
  \caption{Safety vs Passage test for 120$^\circ$ and 360$^\circ$ FoV in
  nonholonomic 1$^{\rm st}$ order case. Performance vs $\rMin$.
  \label{fig:safepass}}
\end{figure}
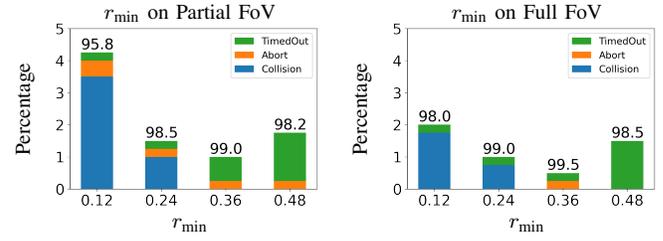

\subsubsection{Nonholonomic, Safety vs Passage}
In \S\ref{sss:nonholo1}, the modified PG method with full FoV did not
achieve perfect navigation as for the PG holonomic case in
\S\ref{sss:holo}. This is a function of the lack of full control, the
tuning of the projection operator, and some non-determinism in the
outcomes.  There is a trade-off between tuning for task completion
versus safety.  Figure \ref{fig:safepass} shows the effect of varying
the parameter $\rMin$ in \eqref{eq:poFun}.  For lower values there are
less aborts or time outs, but at the cost of increased collisions.  For
higher values there are more time outs and some aborts, but no
collisions. The results hold for the two FoV cases tested, which
represent the lowest reasonably safe and available FoV for popular
range-based sensors, and the maximal FoV settings.  Future work aims to
address the trade-off by exploring how to dynamically adjust the
parameter.



\subsubsection{Timing}
The median per trajectory compute time of {\algName} was less than 1ms, while
TEB was just under 10ms, leading to an order of magnitude difference.
The average processing time per frame for PG was 8ms, half that of TEB 
($\sim$17ms). Recall that, on average, PG evaluated 5x more paths
than TEB per invocation. 
%
%



\section{Conclusion}
Potential gap is a local planner, designed using ideas from artificial
potential field theory to create a spatially localized gradient descent
problem with guarantees on safe passage under an idealized robot model
with full sensing. 
Central to the idea is the use of perception space to identify gaps for
constructing the local gradient flow. 
Deviations from this model, such as field of view limitations and
nonholonomic motion models, require modifications of the baseline
method to recover lost safety and performance properties. 
Simulated scenarios demonstrate that the
method achieves high completion rates and matches the safety of an
optimal trajectory synthesis method but with faster runtime and higher
completion rate. The code is open-sourced \cite{ivagitPGap}.

The study is a first step towards provably safe, hierarchical navigation
with realistic robot models and configurations. Additional investigation
is needed to strengthen the safety properties for the nonholonomic
case.  Lastly, more aggressive maneuvering with strong momentum
effects were not considered.  Further study is needed to explore how the
{\algName} method may operate under these conditions.




\bibliographystyle{IEEEtran}
\bibliography{hieNav,gap}

\end{document}